\documentclass[aos]{imsart}
\RequirePackage{amsthm,amsmath,amsfonts,amssymb}
\RequirePackage[numbers]{natbib}
\RequirePackage[colorlinks,citecolor=blue,urlcolor=blue]{hyperref}
\RequirePackage{graphicx}
\usepackage{hyperref}
\usepackage{amssymb}
\usepackage{multirow}
\usepackage{array}
\usepackage{bm}
\usepackage{booktabs}
\usepackage{float}
\usepackage{multirow}
\usepackage{algorithm}
\usepackage{algorithmic}
\usepackage{placeins}
\usepackage{mathtools}
\usepackage{mathrsfs}
\usepackage{dsfont}
\usepackage{relsize}
\usepackage{rotating}
\usepackage{enumitem}
\usepackage{minitoc}
\usepackage{etoc}
\usepackage{listings}

\usepackage{tocloft}
\usepackage[toc,page]{appendix}

\makeatletter
\renewcommand\tableofcontents{%
    \@starttoc{toc}%
}
\makeatother

\startlocaldefs
\theoremstyle{plain}
\newtheorem{thm}{Theorem}
\newtheorem{lemma}{Lemma}
\counterwithin{lemma}{section}
\counterwithin{thm}{section}
\newtheorem{rmk}{Remark}[thm]
\newtheorem{proposition}{Proposition}[thm]

\counterwithin{example}{section}
\counterwithin{rmk}{section}
\counterwithin{proposition}{section}
\newtheorem{corollary}{Corollary}
\counterwithin{corollary}{section}

\counterwithin{assumption}{section}
\newtheorem{definition}{Definition}[thm]
\counterwithin{definition}{section}
\newtheorem{assump}{Assumption}

\DeclareMathOperator*{\argmax}{arg\,max}
\DeclareMathOperator*{\argmin}{arg\,min}
\def\##1\#{\begin{align}#1\end{align}}
\def\$#1\${\begin{align*}#1\end{align*}}

\endlocaldefs

\begin{document}

\begin{frontmatter}
\title{Distributional Shift-Aware Off-Policy Interval Estimation: \\
A Unified Error Quantification Framework}



\runtitle{Distributional Shift-Aware Off-Policy Interval Estimation}

\begin{aug}
\author[A]{\fnms{Wenzhuo    }~\snm{Zhou}\ead[label=e1]{wenzhuz3@uci.edu}},
\author[B]{\fnms{Yuhan}~\snm{Li}\ead[label=e2]{yuhanli8@illinois.edu}}
\author[B]{\fnms{Ruoqing}~\snm{Zhu}\ead[label=e3]{rqzhu@illinois.edu}}
\and
\author[A]{\fnms{Annie}~\snm{Qu}\ead[label=e4]{aqu2@uci.edu}}
\address[A]{University of California Irvine \printead[presep={,\ }]{e1,e4}}

\address[B]{University of Illinois Urbana Champaign\printead[presep={,\ }]{e2,e3}}

\vspace{4mm}

February 15, 2023

\end{aug}

\begin{abstract}

We study high-confidence off-policy evaluation in the context of infinite-horizon Markov decision processes, where the objective is to establish a confidence interval (CI) for the target policy value using only offline data pre-collected from unknown behavior policies. 
This task faces two primary challenges: providing a comprehensive and rigorous error quantification in CI estimation, and addressing the distributional shift that results from discrepancies between the distribution induced by the target policy and the offline data-generating process. Motivated by an innovative unified error analysis, we jointly quantify the two sources of estimation errors: the misspecification error on modeling marginalized importance weights and the statistical uncertainty due to sampling, within a single interval. This unified framework reveals a previously hidden tradeoff between the errors,  which
 undermines the tightness of the CI. Relying on a carefully designed discriminator function, the proposed estimator achieves a dual purpose: breaking the curse of the tradeoff to attain the tightest possible CI, and adapting the CI to ensure robustness against distributional shifts. Our method is applicable to time-dependent data without assuming any weak dependence conditions via leveraging a local supermartingale/martingale structure. Theoretically, we show that our algorithm is sample-efficient, error-robust, and provably convergent even in non-linear function approximation settings. The numerical performance of the proposed method is examined in synthetic datasets and an OhioT1DM mobile health study. 
\end{abstract}



\end{frontmatter}

\section{Introduction}\label{intro}

Off-policy evaluation (OPE) is a crucial task in reinforcement learning (RL). Its primary goal is to evaluate a new policy (known as the target policy) using observational data generated by various existing policies (called behavior policies). 
In many real-world applications, e.g., healthcare \citep{murphy2001marginal,luckett2020estimating}, financial marketing \citep{theocharous2020reinforcement}, robotics \citep{thomas2015high} and education \citep{mandel2014offline}, implementing a new policy can be costly, risky, or unsafe. OPE enables us to evaluate the performance of a new policy using only a set of observational data, without interacting with the real environment. This method substantially reduces the cost and risk associated with deploying a new policy, making it a valuable tool in practice.

There have been many works on point estimations of OPE in recent years. Popular approaches include value-based \citep{le2019batch,liao2020off,chen2022well}, importance sampling based \citep{precup2000eligibility,jiang2016doubly,thomas2016data}, and marginalized importance sampling based approaches \citep{liu2018breaking,xie2019towards,uehara2020minimax, zhang2020gendice,nachum2020reinforcement}. On the theoretical side,
\cite{uehara2020minimax} established asymptotic optimality and efficiency for OPE in the tabular setting, and \cite{kallus2020double} provided a complete study of semiparametric
efficiency in a more general setting. \cite{wang2020statistical} studied the fundamental hardness of OPE with linear
function approximation. We refer readers to \cite{uehara2022review} for a comprehensive review of OPE problems. 

While much of the current research on OPE is centered on point estimation, in many real-world scenarios it is desirable to avoid making overconfident decisions that could result in costly and/or irreversible consequences. That is,
 rather than solely relying on a point estimate, 
many applications would benefit significantly from having a  confidence interval (CI) on the value of a policy.
Existing approaches on off-policy CI estimation are mainly based on asymptotic inference \citep{liao2020off, shi2020statistical,dai2020coindice,shi2022off} or bootstrapping \citep{hanna2017bootstrapping,hao2021bootstrapping,ramprasad2022online}. In particular, \cite{liao2020off} constructed a limiting distribution for the policy value function to estimate CIs in a tabular representation setting, and \cite{shi2020statistical} considered an asymptotic CI for policy value with smoothness assumptions on the action-value function. Typically, the asymptotic inference or bootstrap-based methods require a large number of samples to achieve a desirable coverage. Unfortunately, in practical applications such as mobile health, the number of observations may be limited. For example, the OhioT1DM dataset
\citep{marling2020ohiot1dm}  only contains a few hundred observations \citep{shi2021deeply}. 
In such scenarios, providing non-asymptotic uncertainty quantification to meet practically feasible conditions may be crucial. \cite{thomas2015high} constructed an empirical Bernstein inequality applied to the stepwise importance sampling estimator. However, the estimated CI can become quite loose due to the high variance introduced by the stepwise importance weights, particularly in infinite horizon decision-making settings.  Recently, \cite{feng2020accountable, feng2021non} proposed a finite-sample CI based on a kernel action-value function estimator. However, their approach relies on a strong assumption that the true model must be correctly specified, which may result in significant bias under potential model misspecification. 

More importantly, all the aforementioned CI estimation approaches neglect the influence of the \textit{distributional shift}, which is arguably the most significant challenge in offline RL. The distribution shift refers to the distributional mismatch between offline data distribution and that induced by target policies \citep{levine2020offline}.
Algorithmically, the distributional shift often leads to compounding biased estimations of action values, resulting in unsatisfactory performance for many OPE algorithms \citep{kallus2022doubly}. Practically, in fields such as precision medicine, shifts in treatment regimens may produce arbitrary and sometimes harmful effects that are costly to diagnose and treat  \citep{challen2019artificial}. To mitigate this problem, it is essential to account for the distributional shift when performing high-confidence OPE.    

From the perspective of the error analysis in the off-policy CI estimation, the existing algorithms 
are typically developed based on a separate error quantification. The approaches focus on either quantifying the model misspecification bias, which arises from function approximation while neglecting sampling uncertainty, or on measuring statistical uncertainty under the assumption that the true model is accurately specified. This leads to the true relationship between the ``bias'' and ``uncertainty'' being unrevealed, which in turn creates a gap between algorithm developments and practical applications. How to rigorously quantify the bias and uncertainty in a unified framework is unknown and challenging. Moreover, the existing works usually assume independence or weak dependence, e.g., mixing random processes, in an offline data distribution. This assumption is often violated due to the complicated interdependent samples in many RL real applications \citep{agarwal2019reinforcement}. Additionally, most popular approaches only work when the offline data is induced by a single known or unknown behavior policy. However, they are not applicable in settings involving a mixture of multiple unknown behavior policies \citep{kallus2019intrinsically}.

Motivated by the aforementioned issues,
in this paper we propose a novel, robust, and possibly tightest off-policy CI estimation framework in infinite-horizon Markov decision process (MDP) settings. The advances of the proposed method and our contributions are summarized as follows. First, our proposed framework mitigates the issue of distributional shifts by properly incorporating a carefully designed discriminator function. This discriminator captures information about the distributional shift and adjusts the resulting CI accordingly, alleviating the risk of poor estimates when the distribution of offline data deviates from the distribution under the target policy. By doing so, the estimated CI effectively reduces potential decision-making risks and avoids unsatisfactory performance.

\begin{figure}[h]
        \vspace{-0.5cm}
    \centering
\includegraphics[scale=0.4]{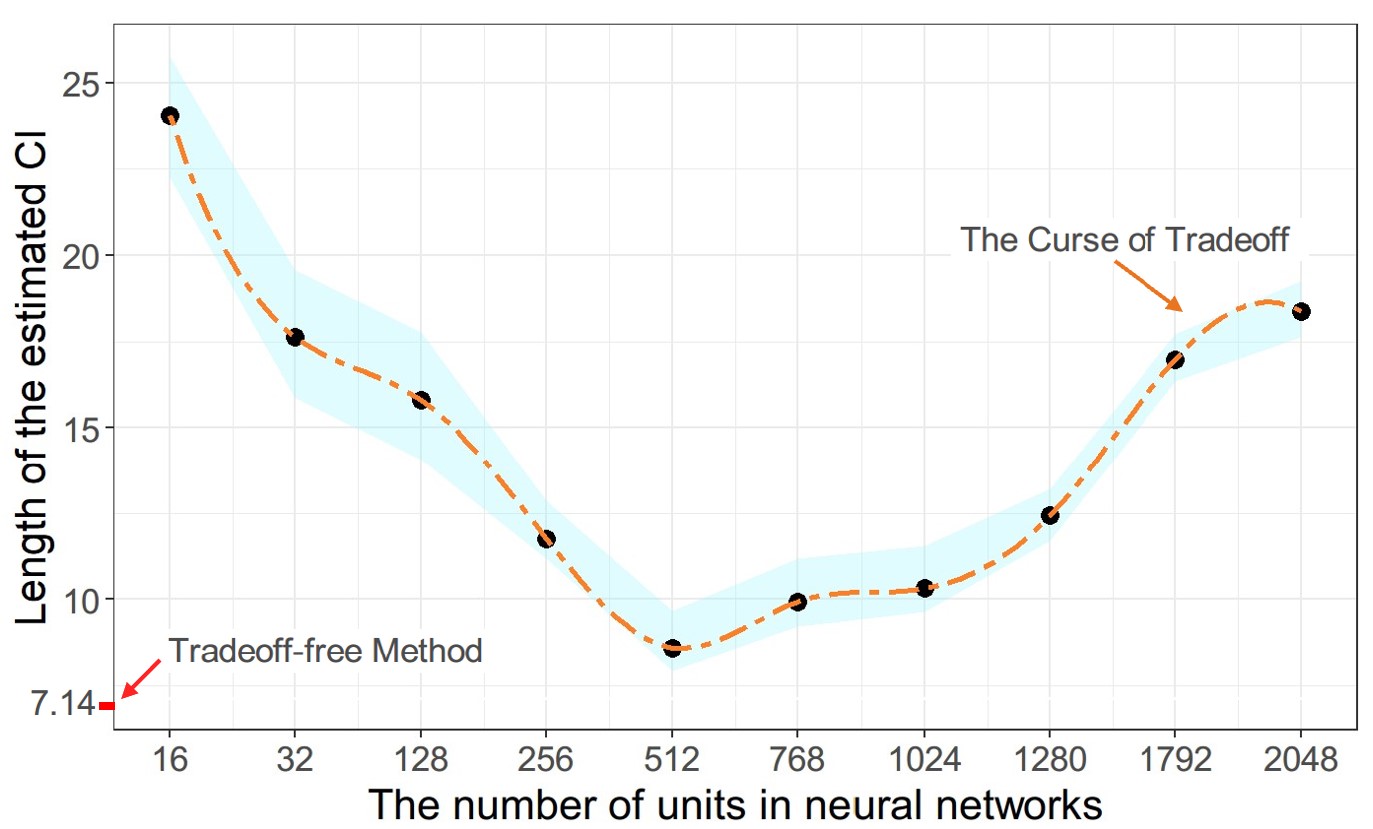}
    \caption{An illustrating example of the curse of the tradeoff in CI estimation from CartPole RL environment. Our unified error quantification analysis shows that the tightness of the estimated CI is determined by both bias and uncertainty, as discussed in Section \ref{bias_uncert}. These factors are influenced by the complexity of the function approximation class. As the model complexity increases (i.e., the number of units increases), the evaluation bias is reduced but the statistical uncertainty increases, and vice versa. This tradeoff makes it challenging to maintain an optimal balance, resulting in a less tight CI. 
    In contrast, our proposed \textit{tradeoff-free} estimation approach effectively breaks this curse, ultimately yielding a tighter CI with a length of $7.14$.}
        \label{fig:tradeoff}
        \vspace{-0.5cm}
\end{figure}

Second, we develop a novel and unified error quantification analysis that reveals a previously hidden tradeoff among the sources of errors. 
 Specifically, our framework allows us to decompose the error in CI estimation into two distinct parts: the evaluation bias induced by a misspecification error of the function
approximator, and the statistical uncertainty due to
sampling. By unifying the two source errors within a single interval, 
we have shown that there is an inherent tradeoff between the errors and can be naturally viewed as a ``bias-uncertianty'' tradeoff, as depicted in Figure \ref{fig:tradeoff}. This tradeoff makes it challenging to strike a balance and obtain the tightest possible CI estimation.

Third, we break the curse of the tradeoff and formulate a \textit{tradeoff-free} framework, resulting in the possibly tightest CI, illustrated in Figure \ref{fig:tradeoff}.
 In smooth MDP settings \citep{shi2020statistical}, we leverage the principle of maximum mean discrepancy to control uncertainty and preserve function expressivity simultaneously. For more general MDPs, we carefully calibrate the uncertainty deviation, making it entirely independent of the model-misspecification bias. We achieve this by leveraging the advantageous curvature of the designed discriminator function. As a result, both evaluation bias and sampling uncertainty can be minimized jointly without compromising one or the other.

Fourth, our method demonstrates broad applicability. To the best of our knowledge, our algorithm is the first general function approximation
algorithm that can be applied to offline data collected without
assuming weakly dependent distribution. The proposed method sufficiently utilizes the (super)martingale structure to handle statistical errors and works particularly efficiently in small sample size settings. Also, our algorithm is stable and efficient for a wide range of function approximators, including deep neural networks. These characteristics significantly expand the application scope of the proposed method.

Lastly, the proposed algorithm is sample-efficient and robust against model-misspecification and optimization errors. By sample efficiency, we mean that the sample complexity requirement for finding a near-tight CI is polynomial with respect to key parameters. We further demonstrate that the finite sample bound for learning a near-tight CI can be independent of the function class complexity, and shrink at a sublinear rate, i.e., $\mathcal{O}(1/n)$, under some mild conditions where $n$ represents the sample size of the offline dataset. 
We make substantial progress towards solving the longstanding open problem in offline RL with non-linear function approximation \citep{mahadevan2014proximal,levine2020offline}. Thanks to the proposed stochastic approximation optimization algorithm, our approach is proven to converge sublinearly to a stationary point with a vanishing gradient even in non-linear (possibly nonconvex) function approximation settings.


The rest of the paper is organized as follows. Section \ref{prelim} lays out the basic model notation
and data-generating process, as well as the connection between the marginalized importance sampling and linear programming. Section \ref{bias_uncert} formally defines the discriminator function and quantifies bias and uncertainty through a unified error quantification analysis. Section \ref{tradeoff_curse} illustrates the curse of the tradeoff between bias and uncertainty, and introduces a novel framework to overcome this challenge.
Section \ref{algo_sec} presents an efficient stochastic approximation algorithm for CI estimation. A comprehensive theoretical study of our algorithm is provided in Section \ref{theoretical_sec}. Sections \ref{simu_sec} and \ref{real_sec} demonstrate the empirical performance of
our method. Section \ref{dis_sec} concludes
this paper with a discussion. All technical proofs and additional discussion can be found in the appendix.

\section{Preliminaries}
\label{prelim}

\subsection{Markov Decision Process}
We consider an infinite-horizon discounted Markov decision process (MDP) $\mathcal{M}=\{\mathcal{S}, \mathcal{A}, \mathds{P}, \gamma, r, s^{0}\}
$,  where $\mathcal{S}$ is the state space, $\mathcal{A}$ is the action space, $
\mathds{P}: \mathcal{S} \times \mathcal{A} \rightarrow \Delta(\mathcal{S})
$ is the Markov transition kernel for some probabilistic simplex $\Delta$, $
r: \mathcal{S} \times \mathcal{A} \rightarrow \mathbb{R}
$ is the reward function, $\gamma \in [0,1)$ is the discounted factor and $s^0$ is the initial state. A policy $\pi: \mathcal{S} \rightarrow  \Delta(\mathcal{A})$  induces a distribution of the trajectory $
s^{0}, a^{0}, r^{0}, s^{1}, \ldots$, where $
a^{t} \sim \pi(\cdot | s^{t}), r^{t}=r(s^{t}, a^{t}), s^{t+1} \sim \mathds{P}(\cdot|s^{t}, a^{t})$ for any $t \geq 0$. The expected discounted return of a policy is defined as $J(\pi) = \mathbb{E}[\sum_{t=0}^{\infty} \gamma^{t} r^{t} | \pi]$. The discounted return when the trajectory starts with $(s, a)$ and all remaining actions are taken according to $\pi$ is called $q$-function $q^{\pi}: \mathcal{S} \times \mathcal{A} \rightarrow (-\infty, \bar{V}]$. The $q^{\pi}$ is the unique fixed point of the Bellman operator $\mathcal{B}^{\pi}
$, satisfying the Bellman equation \citep{sutton2018reinforcement}: 
$\mathcal{B}^{\pi} q(s, a) \coloneqq r(s, a)+\gamma \mathbb{E}_{s^{\prime} \sim \mathds{P}(\cdot | s, a)}[q(s^{\prime}, \pi)].
$
Here $q(s^{\prime}, \pi)$ is denoted as shorthand for $
\mathbb{E}_{a^{\prime} \sim \pi\left(\cdot | s^{\prime}\right)}\left[q\left(s^{\prime}, a^{\prime}\right)\right]
$, and we define $\mathds{P}^{\pi}q(s,a) := \mathbb{E}_{s^{\prime} \sim \mathds{P}(\cdot | s, a)}\left[q\left(s^{\prime}, \pi\right)\right]$. 

Another important notion is 
the normalized discounted visitation of $\pi$, defined as 
$
d^{\pi}(s,a) \coloneqq (1-\gamma)\sum_{t=0}^{\infty} \gamma^{t} d^{\pi,t}(s,a)
$,
where $d^{\pi,t}$ is the marginal state-action distribution at the time-step $t$. Essentially, $d^{\pi}(s,a)$ characterizes the states and
actions visited by $\pi$. For notation convenience,  we write $\mathbb{E}_{d^{\pi}}[\cdot]$ to represent
$
\mathbb{E}_{d^{\pi}}\left[g\left(s, a, r, s^{\prime}\right)\right]:=\int_{s , a} d^{\pi}(s, a) \mathds{1}_{\{r=r(s,a)\}}\linebreak \mathbb{E}_{s^{\prime} \sim \mathds{P}(\cdot|s, a)}\left[g\left(s, a, r, s^{\prime}\right)\right].
$
In the offline RL setting, there exists an unknown offline data-generating distribution $\mu$ induced by behavior policies.  With a slight abuse of notation,  we refer to the marginal distribution over $(s,a)$ and joint distribution over $(s,a,r,s^{\prime})$ as $\mu$. Despite the unknowns of $\mu$, we can observe a set of transition pairs, as offline dataset $\mathcal{D}_{1:n} \coloneqq \{s_i, a_i, r_i, s^{\prime}_i\}^{n}_{i=1}$ sampling from $\mu$. 

We make the following \textit{minimum} data collection condition throughout the paper: 
\$
\mu(r_i,s^{\prime}_{i}|s_i,a_i,s_{i-1},a_{i-1},r_{i-1},s^{\prime}_{i-1},...,s_1,a_1) = \mathds{P}(s^{\prime}_i|s_i,a_i)\mathds{1}_{\{r_i = r(s_i,a_i)\}}.
\$
This condition relaxes the standard assumptions widely made in the RL literature, e.g., \cite{liu2018breaking,shi2020statistical,liao2022batch} typically require the sample path to be either i.i.d or weakly dependent. In contrast, we only require that the reward $r_{i}$ and the transition state $s^{\prime}_{i}$ follow the rules specified by $r$ and $\mathds{P}$, but do not impose any restrictions on the generation of the state-action pair $(s_i,a_i)$. Therefore, the dataset $\mathcal{D}_{1:n}$ can be collected as a set of multiple trajectories with interdependent samples, or under a mix of arbitrary unknown behavior policies.
Ultimately, the goal of high-confidence OPE is to construct an efficient CI for estimating the return of policy $\pi$ using the offline dataset.

\subsection{Marginalized Importance Sampling and Linear Programming}
\label{mis_dis}
In this section, we provide a concise review of two essential concepts in OPE: marginalized importance sampling (MIS) \citep{liu2018breaking} and linear programming (LP) \citep{dai2017boosting}. These concepts form the foundation for our subsequent discussion. Following the definition of the MIS in \cite{liu2018breaking}, we have
$
\tau_{d^{\pi}/ \mu}(s, a):=\frac{d^{\pi}(s, a)}{\mu(s, a)} 
$
to characterize the ratio of the visitation induced by target policy and behavior policy. If it exists, 
$
\mathbb{E}_{d^{\pi}}\left[g\left(s, a, r, s^{\prime}\right)\right]=\mathbb{E}_{\tau_{d^{\pi} / \mu}}\left[g\left(s, a, r, s^{\prime}\right)\right]=\mathbb{E}_{\mu}\left[\tau_{d^{\pi} / \mu}(s, a) g\left(s, a, r, s^{\prime}\right)\right],
$ 
where  $\mathbb{E}_{\tau}[\cdot]:=\mathbb{E}_{\mu}[\tau(s, a)(\cdot)]$ is shorthand used throughout the paper. The return $J(\pi)$ can be obtained via re-weighting the reward with respect to the MIS weight, i.e., 
$
J(\pi)=\frac{\mathbb{E}_{d^{\pi}}[r]}{1-\gamma}=\frac{\mathbb{E}_{\tau_{d^{\pi} / \mu}}[r]}{1-\gamma}
$. Motivated by this fact, \cite{uehara2020minimax,shi2022minimax}  proposed a minimax estimator to estimate $\tau_{d^{\pi} / \mu}$ by searching a weight function $\tau(s,a):= \frac{d(s,a)}{\mu(s,a)}$ in a specified function class $\tau \in \Omega$ approximately:
\#
\min_{\tau \in \Omega} \max_{q \in \mathcal{Q}}W(q,\tau):=\mathbb{E}_{\mu}\left[\gamma \tau(s, a)  q\left(s^{\prime}, \pi\right) -\tau(s, a) q(s, a)\right]+(1-\gamma) q\left(s^0, \pi\right).
\label{mis_estimator}
\#
Once $\tau_{d^{\pi} / \mu}$ is solved, the $d^{\pi}$ can be obtained using the chain rule, i.e., $d^{\pi} = \tau_{d^{\pi} / \mu}\cdot\mu$. 

Recently, \cite{nachum2020reinforcement} showed that the above minimax problem has an equivalent solution to the LP problem \citep{dai2017boosting}: $\text{maximize}_{d} \;\, \mathbb{H}(d) \equiv 0$ constrained on 
\#
& \quad (1-\gamma)\mathds{1}\{s = s^0\} \pi(a | s) +  \gamma \pi(a |s) \sum_{{s}^{\prime}, {a}^{\prime}} \mathds{P}(s|{s}^{\prime}, {a}^{\prime}) d({s}^{\prime}, {a}^{\prime}) = d(s, a) , \; \text{for any} \; s,a,
\label{dual_lp}
\#
 where $\mathbb{H}(\cdot)$ is a functional that takes a visitation $d$ as input and is set to be a constant zero function for the LP problem. It is crucial to note that \eqref{dual_lp} is over-constrained for any given $s, a$. Specifically, the visitation $d^{\pi}$ can be uniquely identified solely based on the constraints. This over-constrained characteristic explains why $\mathbb{H}(d)$ can be set as a zero constant function and why \eqref{dual_lp} is sufficient to recover the optimal solution $d^{\pi}$. In the following, we will utilize this property to incorporate distributional shift information into the proposed framework.

\section{Off-Policy Confidence Interval Estimation}
\label{bias_uncert}

In this section, we propose a novel off-policy CI estimation framework that can adapt to distributional shifts. To develop this framework, we first decompose the source errors in CI estimation into two critical components: evaluation bias, which arises from model misspecifications, and statistical uncertainty stemming from sampling. We subsequently integrate and quantify both components, allowing them to be unified within a single interval. The way to encode the distributional shift information is naturally introduced in evaluation bias quantification.


\subsection{Evaluation Bias Quantification}
\label{bias_quan}

We study the first major component in CI estimation, i.e., evaluation bias, and propose a value interval method to quantify it. The developed framework is robust to model misspecification and naturally encodes the distributional shift information. To begin with, we formally define the discriminator function, which serves a dual purpose in this work. For now, we primarily concentrate on one of these roles, which is to measure the extent of deviation between the offline data-generating distribution $\mu$ and the distribution induced by the target policy. Later, we will demonstrate the other role that the discriminator function plays in estimating CI in Section \ref{tradeoff_curse}.



\begin{definition}
\label{alpha_def}
For $x,c_1,c_2,C \in \mathbb{R}^{+}$ and $C \geq 1$, 
the discriminator function $\mathbb{G}(\cdot)$ satisfies the following conditions: (1) \textbf{Strong convexity}: $\mathbb{G}(x)$ is M-strongly convex with respect to $x$. (2) \textbf{Boundedness}: $|\mathbb{G}(x)| \leq c_1$ for $x \in [0,C]$. (3) \textbf{Boundedness on first-order derivative}: $|\mathbb{G}^{\prime}(x)| \leq c_2$ if $x \in [0,C]$. (4) \textbf{Non-negativity}: $\mathbb{G}(x) \geq 0$. (5) \textbf{1-minimum}: $\mathbb{G}(1) = 0$.
\end{definition}

The family of R\'enyi entropy \citep{renyi1961measures}, Bhattacharyya distance \citep{choi2003feature}, and simple quadratic form functions all satisfy the conditions outlined in Definition \ref{alpha_def}.

\begin{rmk}
   To understand the motivation of the design of the discriminator function, let us consider taking the marginal important ratio $\tau_{d^{\pi}/\mu}(s,a) = d^{\pi}(s,a)/\mu(s,a)$ as the input for $\mathbb{G}(\cdot)$. The $1$-minimum condition $d^{\pi}(s,a)/\mu(s,a)$=1, i.e., $d^{\pi}(s,a)=\mu(s,a)$,   suggests that no distributional shift is present or detected by the discriminator $\mathbb{G}$. The strong convexity property is capable of characterizing the uniqueness of the minimum point and ensuring that the rate of change increases as $d^{\pi}(s,a)$ deviates from 
$\mu(s,a)$. The boundedness condition on the first-order derivative imposes smoothness on $\mathbb{G}$, while the other two conditions render $\mathbb{G}$ practically quantifiable.
\end{rmk}

As discussed in Section \ref{mis_dis}, the LP problem \eqref{dual_lp} is over-constrained and the visitation $d^{\pi}$ can be uniquely identified. Consequently, one may replace the objective $\mathbb{H}(d)$ with another functional, provided that the newly-replaced functional does not affect the optimal solution $d^{*} = d^{\pi}$. Motivated by this observation, we propose to choose
$\mathbb{H}(d) = \lambda\mathbb{E}_{\mu}[\mathbb{G}(\tau(s,a))]$ where $\tau(s,a) := \frac{d(s,a)}{\mu(s,a)}$, and the hyperparameter $\lambda \in \mathbb{R}$ user's preference and degree of sensitivity to the distributional shift. 
When $\lambda < 0$, it indicates that the user favors small distribution shifts and appeals to pessimistic evaluation, which encourages downgrading the policy associated with large distribution shifts. For $\lambda =0$, the user maintains a neutral attitude toward the distributional shift. Furthermore, our framework degenerates to \eqref{dual_lp} or \eqref{mis_estimator} under this condition. From this perspective, both the minimax estimation in \eqref{mis_estimator} and the LP problem in \eqref{dual_lp} can be considered special cases of our approach when we take $\lambda=0$.  According to the connection between \eqref{mis_estimator} and \eqref{dual_lp}, we can convert the newly-defined LP problem with $\mathbb{H}(d) = \lambda\mathbb{E}_{\mu}[\mathbb{G}(\tau(s,a))]$ to a max-min problem. This can be expressed as $\min_{\tau \in \Omega}\max_{q \in \mathcal{Q}}L(q,\tau)$, where 
\$
L(q,\tau) := 
\frac{\mathbb{E}_{\mu}[\tau(s,a)\left(\gamma q\left(s^{\prime}, \pi\right)-q(s, a)\right) + \lambda \mathbb{G}(\tau(s,a))]}{1-\gamma} + q\left(s^0, \pi\right).
\$
Unfortunately, directly optimizing objective function $L(q,\tau)$ has some drawbacks. To comprehensively reveal these drawbacks, we first present a key lemma to support our arguments. 
\begin{lemma}
\label{tele_lemma}
For any target policy $\pi: \mathcal{S} \rightarrow  \Delta(\mathcal{A})$ and  $\tau: \mathcal{S} \times \mathcal{A} \rightarrow \mathbb{R}^{+}$, 
\$
\frac{\mathbb{E}_{\tau}[r(s,a)]}{1-\gamma} - J^{\alpha}(\pi)  = \frac{\mathbb{E}_{\tau}\left[q^{\pi}(s, a)-\gamma q^{\pi}\left(s^{\prime}, \pi\right) + \lambda \mathbb{G}(\tau(s,a))/\tau(s,a)\right]}{1-\gamma} - q^{\pi}(s^0,\pi),
\$
where $J^{\alpha}(\pi) := J(\pi) + \lambda \xi(\mathbb{G},\tau)$ for $\xi(\mathbb{G},\tau) := \mathbb{E}_{\mu}[\frac{ \mathbb{G}(\tau(s,a))}{1-\gamma}]$, and $q^{\pi}$ is the unique fixed point of Bellman equation $\mathcal{B}^{\pi}q=q$. 
\end{lemma}
Lemma \ref{tele_lemma}  suggests that the rationale behind $\min_{\tau\in\Omega}\max_{q\in\mathcal{Q}}L(q,\tau)$ is to find a ``good'' $\tau^{*}$ such that $\frac{\mathbb{E}_{\tau}[r(s,a)]}{1-\gamma}$ and $ J^{\alpha}(\pi)$ are are sufficiently close when $q^{\pi} \in \mathcal{Q}$. Then $\tau^{*}$ can be plugged in $\frac{\mathbb{E}_{\tau}[r(s,a)]}{1-\gamma}$ to approximate $ J^{\alpha}(\pi)$. In fact, this rationale also implicitly requires the existence of $\tau^{*} \in \Omega$. When $\Omega$ is misspecified, the evaluation will exhibit a high bias. For OPE problems significantly affected by the distributional shift, the true model is particularly challenging to specify \citep{jiang2020minimax,chen2022offline}. Additionally, another limitation of the min-max estimation is that the objective function $L(q,\tau)$ completely disregards the reward information. 

Motivated by the above-mentioned drawbacks, we propose a more robust interval method for estimation. Free from any model-specification assumptions on $\Omega$, our proposal implicitly quantifies the evaluation bias resulting from model-misspecification error within the interval. According to Lemma \ref{tele_lemma} and assuming $q^{\pi}$ is correctly specified, the following inequality holds for any $\tau \in \Omega$:
\#
J(\pi) &\geq \inf_{q \in \mathcal{Q}}\left\{ \frac{\mathbb{E}_{\tau}\left[r(s,a)+\gamma q\left(s^{\prime}, \pi\right)-q(s, a) \right]}{1-\gamma} + q(s^0,\pi)\right\} - \lambda^{-} \xi(\mathbb{G},\tau) \notag \\ 
&\coloneqq \inf_{q \in \mathcal{Q}}L(\tau,q,\pi) - \lambda^{-}\xi(\mathbb{G},\tau), 
\label{l_def}
\#
where $\lambda^{-} := -\min\{0,\lambda\}$. 
Similarly, the return $J(\pi)$ from above:
\#
J(\pi) \leq \sup_{q \in \mathcal{Q}}L(\tau,q,\pi) + \lambda^{+}\xi(\mathbb{G},\tau),  \; \text{for any} \; \tau \in \Omega, 
\label{pop_ci_lower}
\#
where $\lambda^{+}:=\max\{0,\lambda\}$. Considering that \eqref{l_def} and \eqref{pop_ci_lower} hold for any $\tau \in \Omega$, we can optimize them over $\tau$ to obtain the tightest possible interval: 
\#
J(\pi) \in \left[\sup_{\tau\in\Omega}\inf_{q \in \mathcal{Q}}L(\tau,q,\pi)- \lambda^{-}\xi(\mathbb{G},\tau),\;  \inf_{\tau \in \Omega}\sup_{q \in \mathcal{Q}}L(\tau,q,\pi) +\lambda^{+}\xi(\mathbb{G},\tau)\right],
\label{lower_upper_true}
\#
where we denote the upper bound and lower bound as $J^{+}(\pi)$ and $J^{-}(\pi)$, respectively. This optimization can also be interpreted as searching for the best $\tau^{*}$ within the class $\Omega$ to minimize the model-misspecification error. Based on the value interval in \eqref{lower_upper_true}, we demonstrate that the evaluation bias, i.e., the model misspecification error of $\Omega$, is implicitly quantified in the interval and can be decomposed into two components.

\begin{thm}[Evaluation bias]
\label{interval_width_bound}
For any target policy $\pi$, the evaluation bias is bounded above, 
\$
J^{+}(\pi) - J^{-}(\pi) \leq 2\bigg(\underbrace{\inf_{\tau \in \Omega}\bigg\{\sup_{q\in \mathcal{Q}}\left|\mathbb{E}_{\tau}\Big[ \frac{\gamma q(s^{\prime},\pi) - q(s,a)}{1-\gamma} + q(s^0,\pi)\Big]\right|}_{\epsilon_{mis}} + \frac{1}{2} \underbrace{|\lambda|\xi(\mathbb{G},\tau)}_{\epsilon_{dist}}\bigg\}\bigg).
\$
\end{thm}
Theorem \ref{interval_width_bound} implies that the evaluation bias consists of two components: the model-misspecification error $\epsilon_{mis}$ and the bias caused by the discriminator function, denoted by $\epsilon_{dist}$.
The latter can be effectively controlled by adjusting the magnitude of $\lambda$. The former, $\epsilon_{mis}$, is dependent on the wellness of the model specification on $\Omega$. In particular, if the class $\Omega$ is well-specified, it is possible for $\epsilon_{mis}$ to diminish to zero. We direct readers to Section {\color{blue}A} in the appendix for a more comprehensive discussion on the tightness and validity of the interval.

\begin{rmk}
In contrast to
the role of  $\Omega$, the class $\mathcal{Q}$ primarily influences bias quantification. However, there is no need to be concerned about a tradeoff between bias and uncertainty since the specification of $\mathcal{Q}$ solely affects bias quantification.  Consequently, an expressive function class can generally be chosen for modeling $\mathcal{Q}$. Furthermore, since the specification of the function class $\mathcal{Q}$ is independent of uncertainty quantification, it is feasible to construct the optimal data-dependent $\mathcal{Q}$ using the information generated by fitting $q^{\pi}$ with any state-of-the-art method, such as double reinforcement learning \citep{kallus2020double}. This may increase the likelihood of correctly specifying $\mathcal{Q}$. We offer an in-depth discussion on the role of $\mathcal{Q}$ and demonstrate that the evaluation bias resulting from the model-misspecification errors of $\mathcal{Q}$ can depend on the expressivity of the feature mapping class in Theorem {\color{blue}B.1}, as provided in the appendix.
\end{rmk}

To conclude this section, we remark that the value interval \eqref{lower_upper_true} directly quantifies the evaluation bias and incorporates a well-controlled adjustment scheme for distributional shifts. As a result, our proposed framework is robust against both model-misspecification errors and bias estimation due to distributional shifts. It is important to note that the value interval \eqref{lower_upper_true} is not a CI, as it does not address the uncertainty arising from sampling. In the following section, we will explore uncertainty quantification using our established interval estimation framework.

\subsection{Statistical Uncertainty Quantification}
\label{uncertain_quan}

In this section, we first examine the potential sources of uncertainty that may appear in CI estimation. We then establish two types of CIs for the scenarios when $\lambda=0$ and $\lambda <0$, and separately quantify the uncertainty in these two CIs. Intuitively, these two CIs represent \textit{neutral} and \textit{pessimistic} attitudes towards distributional shifts. We begin by presenting a key lemma for our proposal. Recall that we \textit{do not impose} any independent or weakly dependent data collection assumptions, e.g., mixing conditions, for uncertainty quantification.
    \begin{lemma}[Empirical evaluation lemma]
\label{emp_tele_lemma}
Given offline data $\mathcal{D}_{1:n}$, for any target policy $\pi$, the following equation holds, 
\#
\bar{J}^{\alpha}(\pi) = \frac{1}{n}\sum^{n}_{i=1}\frac{ \tau(s_i,a_i)(r_i+\gamma \mathds{P}^{\pi}q^{\pi}(s_i,a_i) - q^{\pi}(s_i,a_i)) + \lambda \mathbb{G}(\tau(s_i,a_i))}{1-\gamma} + q^{\pi}(s^0,\pi),
\label{eq:emp_lemma}
\#
where $\bar{J}^{\alpha}(\pi)=J(\pi)+ \lambda \xi_n(\mathbb{G},\tau)$ for $\xi_n(\mathbb{G},\tau) = \frac{1}{n}\sum^{n}_{i=1}\frac{ \mathbb{G}(\tau(s_i,a_i))}{1-\gamma}$.
\end{lemma}

Lemma \ref{emp_tele_lemma} suggests that there are two sources of uncertainty when constructing CI estimation. The first source originates from the estimand   $\{\tau(s_i,a_i)(r_i+\gamma \mathds{P}^{\pi}q^{\pi}(s_i,a_i) - q^{\pi}(s_i,a_i))\}^{n}_{i=1}$ and can be approximated by  $\{\tau(s_i,a_i)(r_i+\gamma q^{\pi}(s^{\prime}_i,\pi) - q^{\pi}(s_i,a_i))\}^{n}_{i=1}$. The second source stems from the discriminator $\{\mathbb{G}(\tau(s_i,a_i))\}^{n}_{i=1}$. Once the uncertainty of these two sources is quantified, it can be incorporated into the CIs according to Lemma \ref{emp_tele_lemma}.

\subsection{Uncertainty in Neutral Confidence Interval}
In the following, we present our results on uncertainty quantification and establish the CI for the case when $\lambda=0$, which implies that distributional shift is not taken into account.  Specifically, we characterize the uncertainty deviation, denoted by $\sigma_n$ in \eqref{prob_cond_ci}, which constitutes a major component of the proposed \textit{neutral} CI in \eqref{noreg_ci}.

\begin{thm}[Neutral CI]\label{thm_noreg}
For any target policy $\pi$, we suppose $q^{\pi}\in \mathcal{Q}$ and $\tau \in \Omega$, then the return $J(\pi)$ falls into the following CI with at least probability $1-\delta$,
\#
J(\pi) \in \left[\frac{1}{n}\sum^{n}_{i=1}\frac{r_i\tau(s_i,a_i)}{1-\gamma} - \sup_{q \in \mathcal{Q}}\widehat{M}^{\tau}_{n}(-q,\tau) - \sigma_n, \frac{1}{n}\sum^{n}_{i=1}\frac{r_i\tau(s_i,a_i)}{1-\gamma} + \sup_{q \in \mathcal{Q}}\widehat{M}^{\tau}_{n}(q,\tau) + \sigma_n \right],
\label{noreg_ci}
\#
if the uncertainty deviation $\sigma_n$ satisfies
\#
P\left(\sup_{\tau \in \Omega}\Big| \widehat{L}^{q}_n(q^{\pi},\tau) \Big|\leq \sigma_n \right) \geq 1-\delta,
\label{prob_cond_ci}
\#
where 
\$
\widehat{L}^{q}_n(q,\tau) &:= \frac{1}{n}\sum^{n}_{i=1} \frac{\tau(s_i,a_i)\left(q(s_i,a_i)-r_i -\gamma q(s^{\prime}_i,\pi)\right)}{1-\gamma},\\
\widehat{M}^{\tau}_{n}(q,\tau) &:= \frac{1}{n}\sum^{n}_{i=1}\frac{ \tau(s_i,a_i)(\gamma q(s^{\prime}_i,\pi) - q(s_i,a_i))}{1-\gamma} + q(s^0,\pi).
\$
\end{thm}
A closer examination of \eqref{noreg_ci} and \eqref{prob_cond_ci} reveals two key insights. Regarding the class $\mathcal{Q}$, even though the bias quantification necessitates searching possible $q$ uniformly over $\mathcal{Q}$, the statistical uncertainty  deviation $\sigma_n$ can be evaluated point-wisely, specifically at the true $q$-function $q^{\pi}$, as shown in \eqref{prob_cond_ci}. 
Regarding the class $\Omega$, the result in \eqref{noreg_ci} holds for any $\tau \in \Omega$, even if the specified $\Omega$ class does not capture the true $\tau$, i.e., $\tau_{d^{\pi}/\mu}$. This suggests that the CI in \eqref{noreg_ci} is robust against model misspecification errors. The misspecification of $\Omega$ only influences the tightness of the CI but does not compromise the validity of the CI. Also, optimizing the confidence lower and upper bounds in \eqref{noreg_ci} over $\tau \in \Omega$ can help tighten the CI.

We still need to determine the uncertainty deviation $\sigma_n$ in \eqref{prob_cond_ci} to obtain a tractable CI. Typically, the deviation $\sigma_n$ depends on the complexity of hypothesis function class $\Omega$ and uniform  concentration arguments. To measure the function class complexity, we use a generalized notation for the VC-dimension, called the \textit{pseudo-dimension} (see Definition {\color{blue}C.3} in the appendix). As for the uniform concentration arguments, we construct a martingale difference sequence, which allows us to apply the empirical Freedman's concentration inequality \citep{tropp2011freedman}.  In the following, we illustrate our general result for determining $\sigma_n$ using $L_{\infty}$ bounded classes for $\Omega$ modeling.

\begin{thm}[$L_{\infty}$-function class]
\label{general_func_noreg}

For any target policy $\pi$, we define a composite function class $\mathcal{G} \circ \Omega= \{ g \circ \tau: g \in \mathcal{G}, \tau \in \Omega\}$ such that $
g \circ \tau (s,a,r,s^{\prime}) = {\tau}(s,a)\big(r(s,a)+\gamma {q}^{\pi} (s^{\prime},{\pi})-{q}^{\pi}(s, a)\big),
$
for any $(s,a,r,s^{\prime})$. Suppose $0\leq \tau(s,a)\leq C$ for any $s,a$, then Theorem \ref{thm_noreg} holds if we choose 
\#
\sigma_n =  C\sqrt{\frac{2 \bar{V}^2\ln\frac{8\mathcal{N}\left(1/\sqrt{n}, \mathcal{G} \circ \Omega, \|\cdot\|_{L_2}\right)}{\delta}}{n(1-\gamma)^2}} + \frac{2 C \bar{V} \ln\frac{8\mathcal{N}\left(1/\sqrt{n}, \mathcal{G} \circ \Omega, \|\cdot\|_{L_2}\right)}{\delta}}{3(1-\gamma)n},
\label{dev}
\#
for 
$
\mathcal{N}\left(\epsilon, \mathcal{G} \circ \Omega, \|\cdot\|_{L_2}\right) \leq (e\left(D_{\Omega}+1\right)(4 e \bar{V}^2)^{D_{\Omega}})\left({1}/{\epsilon}\right)^{D_{\Omega}}
$,
where $\mathcal{N}\left(\epsilon, \mathcal{G} \circ \Omega, \|\cdot\|_{L_2}\right)$ is the $\epsilon$-covering number (see Definition {\color{blue}C.1} in the appendix) for $\mathcal{G} \circ \Omega$ , and $D_{\Omega}$ is the pseudo dimension of the function class $\Omega$. 
\end{thm}

As shown in \eqref{dev}, the deviation $\sigma_n$ increases with the pseudo-dimension of $\Omega$ and decreases with the sample complexity of order $\mathcal{O}(1/\sqrt{n})$. Note that this is an information-theoretical result as finding the exact value of the pseudo-dimension for an arbitrary $L_{\infty}$-class function is challenging. To refine our result, we consider using a feature mapping class for modeling $\Omega$. Let $\phi: \mathcal{S} \times \mathcal{A} \rightarrow \mathbb{R}^d$ be a $d$-dimensional feature mapping, and define a feature mapping class $\Omega_{\psi}$ as follows:
$
\Omega_{\psi} := \{(s, a) \mapsto\langle\phi(s, a), \psi\rangle \,| \, \|\psi\|_{L_2} \leq\text{diam}_{\psi}\}
$
where $\text{diam}_{\psi} \in (0, \infty)$. 
\begin{corollary}
\label{linear_class_noreg}
For any target policy $\pi$, we suppose $\tau \in \Omega_{\psi}$, then Theorem \ref{thm_noreg} holds if we choose 
\#
\sigma_n =  \frac{\bar{V}\sqrt{2\lambda_{\max}(\Sigma_n) \ln\frac{4d\sqrt{n}\bar{V}\text{diam}_{\psi}+2}{\delta}}}{2(1-\gamma)n} + \frac{2\text{diam}_{\psi}\bar{V} \ln\frac{\left(4d\sqrt{n}\bar{V}\text{diam}_{\psi}+2\right)}{\delta}}{3(1-\gamma) n},
\label{lr_sigma_n}
\#
where $\lambda_{\max}(\Sigma_n)$ is the max eigenvalue of the matrix $\Sigma_n  = \sum^{n}_{i=1}\{\phi(s_i,a_i)\phi^{\top}(s_i,a_i)\}$. 
\end{corollary}

Corollary \ref{linear_class_noreg} provides a tractable quantification of the uncertainty deviation $\sigma_n$. Each element in \eqref{lr_sigma_n} is either user-specified or can be calculated using the offline data $\mathcal{D}_{1:n}$. Compared to the vanilla rate of $\mathcal{O}(1/\sqrt{n})$ in Theorem \ref{general_func_noreg}, the uncertainty deviation can vanish at a faster sublinear rate, i.e., $\mathcal{O}(1/n)$ if $\lambda_{\max}(\Sigma_n) \ll \text{diam}_{\psi}$. This improvement in the rate of convergence benefits from our careful analysis of variance terms w.r.t. $\Omega_{\psi}$. 

\subsection{Uncertainty in Pessimistic Confidence Interval}
In this section, we establish the CI for $\lambda < 0$, which we call the \textit{pessimistic} CI. Since $\lambda \neq 0$, the pessimistic CI incorporates a newly-defined discriminator function in \eqref{alpha_def}. In the case of $\lambda < 0$, the \textit{pessimistic} CI favors target policies with small distributional shifts. This property allows the \textit{pessimistic} CI to be used to differentiate the goodness of policies that suffer from bias estimation issues due to distributional shifts \citep{levine2020offline}. Additionally, compared to the uncertainty quantification in the \textit{neutral} CI, we need to take one more step to quantify the uncertainty arising from the estimand of the discriminator function. In the following theorem, we present our formal result in statistical sampling uncertainty quantification and establish the \textit{pessimistic} CI.

\begin{thm}[Pessimistic CI]\label{reg_ci_thm}
For any target policy $\pi$, we suppose $q^{\pi}\in \mathcal{Q}$, then the return $J(\pi)$ falls into the CI for any $\tau \in \Omega$ with probability at least $1-\delta$, 
\#
\bigg[\frac{1}{n}\sum^{n}_{i=1}\frac{r_i\tau(s_i,a_i)}{1-\gamma} - \sup_{q \in \mathcal{Q}}\widehat{M}^{\tau}_{n}(-q,\tau) & -  \lambda^{-}\xi_n(\mathbb{G},\tau)  - \sigma^{L}_n, \notag \\
& \frac{1}{n}\sum^{n}_{i=1}\frac{r_i\tau(s_i,a_i)}{1-\gamma} + \sup_{q \in \mathcal{Q}}\widehat{M}^{\tau}_{n}(q,\tau) + \sigma^{U}_n\bigg],
\label{reg_ci}
\#
if the uncertainty deviations $\sigma^{L}_n$ and $\sigma^{U}_n$ satisfy
\#
P\Bigg(\left\{\sup_{\tau \in \Omega} \widehat{L}^{q}_n(q^{\pi},\tau) \leq  \sigma^{U}_n\right\} \;\bigcap\; \left\{\inf_{\tau \in \Omega}\big\{ \widehat{L}^{q}_n(q^{\pi},\tau) +\lambda^{-}\xi_n(\mathbb{G},\tau)\big\} \geq -\sigma^{L}_n \right\}\Bigg) \geq 1-\delta.
\label{reg_prob}
\#
\end{thm}
The confidence upper bound in \eqref{reg_ci} is nearly identical to the one in the \textit{neutral} CI, so we will not discuss it further here. However, the confidence lower bound in \eqref{reg_ci} differs from the one in \eqref{noreg_ci}. The current confidence lower bound incorporates distributional shift information as well as the corresponding uncertainty included in the deviation $\sigma^{L}_{n}$. By doing so, the estimand of the discriminator $\xi_n(\mathbb{G},\tau)$ helps to offset the bias evaluation caused by distributional shifts.
When users follow the pessimism principle \citep{kumar2019stabilizing,jin2021pessimism}, the confidence lower bound can make policy evaluation and optimization more reliable when the offline dataset is poorly explored or exhibits significant distributional shifts.  
Careful readers may notice that the involvement of the discriminator tends to widen the interval. However, the degree of widening can be well-controlled by $\lambda^{-}$. Additionally, due to the potentially tightest construction of the CI, which will be demonstrated in the next section, the \textit{pessimistic} CI does not lead to overly pessimistic reasoning even with the inclusion of an extra discriminator.

To make \eqref{reg_ci} estimable, we need to determine the uncertainty deviations, $\sigma^{U}_n$ and $\sigma^{L}_n$. For $\sigma^{U}_n$, we can easily follow the derivation in the \textit{neutral} CI. However, for $\sigma^{L}_n$, the quantification is slightly more complicated. To determine  $\sigma^{L}_n$, we construct a pseudo estimator 
\$
\mathcal{U}(s_i,a_i) :=(1-\gamma)^{-1} \mathbb{E}_{s^{\prime} \sim \mathds{P}(\cdot|s_i,a_i)}[\tau(s_i,a_i)(q(s_i,a_i)-r_i -\gamma q(s^{\prime},\pi)) + \lambda^{-}\mathbb{G}(\tau(s_i,a_i))],
\$
such that $\{\widehat{L}^{q}_n(q^{\pi},\tau) +\lambda^{-}\xi_n(\mathbb{G},\tau)- n^{-1}\sum^{n}_{i=1}\mathcal{U}(s_i,a_i)\}$ forms a summation of a martingale difference sequence. We can then follow an almost identical martingale concentration  technique as in Theorem \ref{general_func_noreg} to determine $\sigma^{L}_{n}$, so we omit it here. In the next section, we will provide a more elegant and tighter approach for quantifying the uncertainty deviation $\sigma^{L}_{n}$.

\section{Breaking the Curse of Tradeoff}
\label{tradeoff_curse}

In this section, we first highlight a finding that there exists a hidden tradeoff between the evaluation bias, resulting from model-misspecification, and the statistical uncertainty originating from sampling, as implied by the unified error quantification analysis for CI estimation. This \textit{seeming} conflict prevents the simultaneous minimization of these two sources of errors in order to obtain tighter CIs. This phenomenon can be naturally viewed as the curse of the ``bias-uncertainty'' tradeoff, as shown in Figure \ref{fig:tradeoff}. On this observation, we propose two efficient solutions to break the curse for the \textit{neutral} and \textit{pessimistic} CI, respectively.

\subsection{Bias and Uncertainty Tradeoff}

As discussed in the previous sections, bias and uncertainty are two major sources of errors in CI estimations. In order to obtain a tight interval, one would expect to minimize both of them. 
Statistical uncertainty, as shown in \eqref{dev}, is highly dependent on the complexity of the function class, which is measured in terms of the pseudo-dimension $D_{\Omega}$. In general, a high pseudo-dimension indicates that the set of functions is complex and thus requires a sufficiently large deviation to control the uncertainty uniformly over all functions in the set. To minimize the uncertainty deviation, we prefer to model $\Omega$ using a parsimonious class of functions with a low pseudo-dimension. As for evaluation bias, Theorem \ref{interval_width_bound} indicates that the model-misspecification error of $\Omega$ is a major component. Minimizing bias requires minimizing the model-misspecification error. Generally, a more expressive function class, i.e., with a high pseudo-dimension, is more likely to capture the true model and potentially reduce the bias due to model misspecification. For example, \cite{shi2022off} allow the VC-dimension to diverge with the sample size to reduce the estimator's bias due to model misspecification. If we translate this principle to our framework, it is equivalent to allowing the pseudo-dimension $D_{\Omega}=\mathcal{O}(n^{-\kappa})$, where $0<\kappa \leq 1$.

Taking both bias and uncertainty into account, we regrettably identify a tradeoff between them, as exemplified in Figure \ref{fig:tradeoff}. On one hand, a more parsimonious function class can attain enhanced efficiency in uncertainty quantification. However, this comes at the expense of sacrificing the function class's expressivity, consequently increasing the bias. On the other hand, a more expressive function class aids in reducing model misspecification errors but demands a higher cost to account for the uncertainty inherent in a complex set of functions. 
\textit{Is there a way to break the curse of the bias and uncertainty tradeoff?} In the following sections, we will present our solution to this question.



\subsection{Tradeoff-Free Neutral Confidence Interval}

In this section, we propose a kernel representation approach to break the curse of the tradeoff for the \textit{neutral} CI in \textit{smooth} MDP settings. Under this representation, the uncertainty deviation is independent of the function class complexity. We make use of a powerful tool in concentration measures, Pinelis' inequality \citep{pinelis2012optimum}, to quantify the data-dependent uncertainty deviation in Hilbert space. As a result, this can achieve minimum costs regarding uncertainty quantification, while simultaneously allowing the model-misspecification error to be arbitrarily minimized due to the flexibility of the kernel representation. To begin with, we introduce the definition of the smooth MDP. 

\begin{definition}
Given an infinite-horizon discounted MDP $\mathcal{M}=\left\{\mathcal{S}, \mathcal{A}, \mathds{P}, \gamma, r, s^{0}\right\}
$, suppose the reward function $r(s,a)$ and the Markov transition kernel $\mathds{P}(\cdot|s,a)$ are continuous over $(s,a)$, then we call $\mathcal{M}$ the \textit{smooth} MDP and denote it as $\widetilde{\mathcal{M}}$.   
\end{definition}

Smooth MDP settings are widely satisfied in many real-world applications and studies in the existing literature \citep{li2022reinforcement,shi2020statistical}. Recall that our goal is to quantify the uncertainty deviation
 $\sup_{\tau \in \Omega}| \widehat{L}^{q}_n(q^{\pi},\tau)|$ in \eqref{prob_cond_ci} to establish the CI. According to theorem 3.2 in \cite{zhou2022estimating}, the maximizer of the problem 
$\sup_{\tau \in \Omega}| \widehat{L}^{q}_n(q^{\pi},\tau)|$, i.e., $\tau^{*}(s,a)$, is a continuous function in the smooth MDP setting $\mathcal{M} = \widetilde{\mathcal{M}}$. This motivates us to consider some smooth function classes for modeling $\Omega$, provided that these function classes can approximate continuous functionals sufficiently well. In particular, we model $\Omega$ in a bounded reproducing kernel Hilbert space (RKHS) equipped with a positive definite kernel $K(\cdot, \cdot)$, i.e.,  $\Omega_{RKHS}(C_K) \coloneqq \{\tau \in RKHS: \| \tau \|_{RKHS} \leq C_K \}$, where $\|\cdot\|_{RKHS}$ denotes the kernel norm and the constant $C_K >0$. This kernel representation allows the maximization problem $\sup_{\tau \in \Omega_{RKHS}(C_K)}| \widehat{L}^{q}_n(q^{\pi},\tau)|$ to have a simple closed-form solution $\tau^{*}$. We note that this is based on the maximum mean discrepancy \citep{gretton2012kernel}. Accordingly, the maximum value $(1-\gamma)\widehat{L}^{q}_n(q^{\pi},\tau^{*})$ equals to 
\#
\sqrt{\frac{C_K}{n^2}\sum_{ij}({q}^{\pi}(s_i, a_i) - r_i - \gamma {q}^{\pi} (s^{\prime}_i,{\pi}))K(\{s_i,a_i\},\{s_j,a_j\})({q}^{\pi}(s_j, a_j) - r_i - \gamma {q}^{\pi} (s^{\prime}_j,{\pi}))},
\label{kernel_loss}
\#
where the square root operator is well-defined by the positive definite kernel $K(\cdot, \cdot)$.

The closed-form solution assists us to elaborate on uncertainty quantification. In \eqref{kernel_loss}, the function $\tau$ is maximized out and no longer appears in the sample estimand. Remarkably, the complexity of $\Omega$ does not influence the uncertainty deviation, which is now entirely independent of the function class complexity. This transforms the uniform uncertainty quantification into a point-wise uncertainty quantification focused on the global maximizer $\tau^{*}$, significantly reducing the uncertainty deviation. Furthermore, it has been demonstrated that Reproducing Kernel Hilbert Spaces (RKHS) can achieve arbitrarily small approximation errors when approximating continuous functions \citep{bach2017breaking}. In other words, it is possible to minimize the model-misspecification error of $\Omega$ and even eliminate the error entirely, as long as the true model $\tau^{\star}$ is continuous, which holds in smooth MDP settings.

In summary, by leveraging the kernel representation, we observe a \textit{double descent} phenomenon whereby the model-misspecification error and uncertainty deviation can be minimized simultaneously. Imposing a minor constraint on the continuity of MDPs, we overcome the curse of the bias and uncertainty tradeoff.  In the following, we provide a brief overview of our method for quantifying the uncertainty deviation under the kernel representation.
Inspired by the reproducing property, we initially construct a pseudo-estimator, i.e., \$
\Lambda^{\text{pseudo}}(\cdot;i) := \langle \mathbb{E}_{s^{\prime}\sim \mathds{P}(\cdot|s_i,a_i)}[{q}^{\pi}(s_i, a_i) - r(s_i,a_i) - \gamma {q}^{\pi} (s^{\prime},{\pi})], \sqrt{C_K}K(\{s_i,a_i\},\cdot) \rangle,
 \$ 
 and rewrite \eqref{kernel_loss} with
  $
\widehat{L}^{q}_n(q^{\pi},\tau^{*}) = \sqrt{\|\frac{1}{n}\sum^{n}_{i=1}\Lambda(\cdot;i)\|_{RKHS}},
 $
where 
 $
\Lambda(\cdot;i) := \langle {q}^{\pi}(s_i, a_i) - r_i - \gamma {q}^{\pi} (s^{\prime},{\pi}),  \sqrt{C_K}K(\{s_i,a_i\},\cdot) \rangle.
 $
It can be seen that the sequence $\{\Lambda(\cdot;i) - \Lambda^{\text{pseudo}}(\cdot;i)\}^{n}_{i=1}$ forms a martingale difference sequence in Hilbert space with respect to the filtration $\mathcal{F}_{i} := (s_{\leq i},a_{\leq i},r_{\leq i-1},s^{\prime}_{\leq i-1})$. 
This motivates us to use Pinelis' inequality to quantify the uncertainty deviation as illustrated in the following theorem. 

\begin{thm}
\label{rkhs_concen}
Given the offline data $\mathcal{D}_{1:n}$, we suppose the kernel $\|K(\cdot,\cdot)\|_{\infty} \leq \bar{K}$, then
\$
\sqrt{\Big\|\frac{1}{n}\sum^{n}_{i=1}\Lambda(\cdot;i)\Big\|_{RKHS}} \leq \frac{\bar{V}\sqrt{\bar{K}C_K }}{2}\sqrt{\frac{2 \ln \left(2/\delta\right)}{n(1-\gamma)}} + \frac{2 \ln \left(2/\delta\right) \bar{V}\sqrt{\bar{K}C_K }}{3(1-\gamma)n} := \varepsilon^{K}_n,
\$
holds with probability at least $1-\delta$. 
Furthermore, Theorem \ref{thm_noreg} holds for any $\tau \in \Omega_{RKHS}(C_K)$ when it takes $\sigma_n = \varepsilon^{K}_n$. 
\end{thm}

Theorem \ref{rkhs_concen} achieves an order of $\mathcal{O}(1/\sqrt{n})$ convergence rate, which is as fast as the minimax rate in learning with i.i.d. data \citep{uehara2022finite}. Enabled by the kernel representation, the upper bound $\varepsilon^{K}_n$ is independent of the complexity of $\Omega$. Let $\sigma_n = \varepsilon^{K}_n$, then the uncertainty deviation is significantly reduced compared to those in 
 Theorem \ref{general_func_noreg} or Corollary \ref{linear_class_noreg}. Recall that we can optimize the upper and lower bound in the \textit{neutral} CI \eqref{noreg_ci} to minimize the bias, and obtain a possibly tightest interval. Intriguingly, the maximization problem $\sup_{q \in \mathcal{Q}}\widehat{M}^{\tau}_{n}(-q,\tau)$ within the CI also has a simple closed-form solution when $\mathcal{Q}$ belongs to a bounded RKHS \citep{liu2018breaking}. This can be employed to alleviate the optimization burden and enhance stability. Now, together with the uncertainty defined in Theorem \ref{rkhs_concen}, we formulate the \textit{trade-off} neutral CI as follows. 

\begin{corollary}[RKHS reformulation]
\label{rkhs_tightest_coro}
On the conditions of Theorem \ref{thm_noreg}, suppose $\Omega$ and $\mathcal{Q}$ are in a bounded RKHS with radius $1$, and we set $\sigma_n =\varepsilon^{K}_n$ with $C_K=1$ in Theorem \ref{rkhs_concen}, then the return $J(\pi)$ 
 falls into the CI   
\#
\left[\sup_{\tau\in\Omega}\left\{\frac{1}{n}\sum^{n}_{i=1}\frac{r_i\tau(s_i,a_i)}{1-\gamma}-\widehat{M}^{K}_{n}(\pi)- \sigma_n\right\},\; \inf_{\tau\in\Omega}\left\{\frac{1}{n}\sum^{n}_{i=1}\frac{r_i\tau(s_i,a_i)}{1-\gamma}+\widehat{M}^{K}_{n}(\pi)+\sigma_n\right\} \right],
\label{double_kernel_ci}
\#
with probability at least $1-\delta$. The detailed expression of $\widehat{M}^{K}_{n}(\tau)$ and derivations are provided in Section {\color{blue}E.11} of the appendix. 
\end{corollary}

As demonstrated in Corollary \ref{rkhs_tightest_coro}, the unstable two-stage optimization problem, involving optimizing both $\tau$ and $q$, is decoupled into an easier solvable single-stage optimization. 


\begin{rmk}
Finding global optimizers in \eqref{double_kernel_ci} is quite challenging due to the need to solve an infinite-dimensional problem in RKHS. However, approximate solvers can be employed to replace the global optimizer in \eqref{double_kernel_ci}, which is typically difficult to find. Thanks to the robustness property of our CI with respect to the model-misspecification error, using approximate solvers only makes the interval slightly wider but does not compromise the validity of the CI. Consequently, it is feasible to utilize random feature approximation \citep{rahimi2007random} or the Riesz representation theorem to transform a complex infinite-dimensional optimization problem into an easily solvable finite-dimensional optimization problem.
\end{rmk}

\subsection{Tradeoff-Free Pessimistic Confidence Interval}

In this section, we introduce a tradeoff-free \textit{pessimistic} CI estimation. We make a further step in breaking the curse of the ``bias-uncertainty'' tradeoff \textit{without} requiring any additional conditions, e.g., smoothness conditions for MDPs as discussed in the last section. In any general MDPs, the evaluation bias and statistical uncertainty in CI estimation can be minimized simultaneously without compromising one or the other. The main idea is to utilize Lagrangian duality and the curvature of the discriminator function $\mathbb{G}$. This is another motivation for defining a discriminator function as in \eqref{alpha_def}, in addition to its role in handling distributional shifts. In the following, we characterize a unique and global optimizer related to the lower bound in \eqref{reg_prob}, which builds a foundation for our subsequent methodological developments.

\begin{proposition}\label{closed_tau}
For any $s,a$, the global optimizer of the maximization problem 
\#
\max_{\tau\geq0} \frac{\tau(s,a)}{1-\gamma}\big(r(s,a)+\gamma\mathbb{E}_{s^{\prime}\sim \mathds{P}(\cdot|s,a)}[q(s^{\prime},\pi)]-q(s,a)\big) - \frac{\lambda^{-}\mathbb{G}(\tau(s,a))}{1-\gamma},
\label{pseudo_max}
\#
is \textbf{unique}. Moreover, it has the following analytical form: 
\#
\tau^{\star}(s,a;q) :=  \left[\left(\mathbb{G}^{\prime}\right)^{-1}\left(\frac{r(s,a)+\gamma \mathbb E_{s^{\prime} \sim \mathds{P}(\cdot|s,a)}[q(s^{\prime},\pi)] - q(s,a)}{\lambda^{-}}\right)\right]^{+},
\label{ana_form}
\#
where $\left(\mathbb{G}^{\prime}\right)^{-1}$ is the inverse function of the derivative $\mathbb{G}^{\prime}$, which is strictly increasing due to the strict convexity of $\mathbb{G}$. 
\end{proposition}

The proof of Proposition \ref{closed_tau} relies on the Lagrangian duality and the strong convexity of $\mathbb{G}$. The analytical form \eqref{ana_form} motivates us to construct an appropriate pseudo estimator to measure the uncertainty deviation which is independent of the complexity of $\Omega$. Define the pseudo estimator  $\Gamma^{\text{pseudo}}$ as
\$
\frac{\sum^{n}_{i=1} \tau^{\star}(s_i,a_i;q^{\pi})\big(r(s_i,a_i)+\gamma\mathbb{E}_{s^{\prime}\sim \mathds{P}(\cdot|s_i,a_i)}[q^{\pi}(s^{\prime},\pi)]-q^{\pi}(s_i,a_i)\big) - \lambda^{-}\mathbb{G}(\tau^{\star}(s_i,a_i;q^{\pi}))}{n(1-\gamma)}, 
\$ 
which can be regarded as the post-optimization objective function of \eqref{pseudo_max}. Correspondingly, we can define its transition sample counterpart, where the unknown Markov transition kernel $\mathds{P}$ is replaced by a transition pair, as 
\$
\widetilde{\Gamma} = \frac{\sum^{n}_{i=1} \widetilde{\tau}^{\star}(s_i,a_i,r_i,s^{\prime}_i;q^{\pi})\big(r_i+\gamma q^{\pi}(s^{\prime}_i,\pi)-q^{\pi}(s_i,a_i)\big) - \lambda^{-}\mathbb{G}(\widetilde{\tau}^{\star}(s_i,a_i,r_i,s^{\prime}_i;q^{\pi}))}{n(1-\gamma)}, 
\$
where $\widetilde{\tau}^{\star}(s_i,a_i,r_i,s^{\prime}_i;q^{\pi})$ is the transition pair counterpart of $\tau^{\star}(s_i,a_i;q^{\pi})$,  accordingly. By Proposition \ref{closed_tau}, it can be seen that the objective function  $-\widehat{L}^{q}_n(q^{\pi},\tau) -\lambda^{-}\xi_n(\mathbb{G},\tau)$ in confidence set \eqref{reg_prob} is bounded above by $\widetilde{\Gamma}$ for any $\tau \in \Omega$. With this fact, we have 
\#
\sup_{\tau \in \Omega}\{-\widehat{L}^{q}_n(q^{\pi},\tau) -\lambda^{-}\xi_n(\mathbb{G},\tau)\} \leq 
 \widetilde{\Gamma} \overset{\mathrm{(*)}}{=} \widetilde{\Gamma} 
 - \Gamma^{\text{pseudo}}.
 \label{transfer}
\#
The equality $\mathrm{(*)}$ comes from the fact that $\Gamma^{\text{pseudo}}=0$ by the Bellman equation.
In \eqref{transfer}, the uniform quantification argument on the LHS is transferred to a point-wise deviation on the RHS. This indicates that the uncertainty deviation 
is sufficient to only be evaluated at the point of the global optimizer $\tau^{\star}$, which removes the influence of the complexity of function class $\Omega$ on the uncertainty quantification. Next, we need to find the uncertainty deviation $\widetilde{\Gamma}  - \Gamma^{\text{pseudo}}$. However, the martingale structure we utilized to establish the uncertainty deviation no longer applies to the difference sequence $\{\widetilde{\Gamma}_i -  \Gamma^{\text{pseudo}}_i\}^{n}_{i=1}$, due to the non-linearity of $(\mathbb{G}^{\prime})^{-1}$ and the \textit{double sampling} issue \citep{baird1995residual} in $\widetilde{\Gamma}_i$. This makes derivation much more intricate compared to the previous sections. Interestingly, by exploiting the curvature, i.e., the second-order derivative of $\Gamma^{\text{pseudo}}_i$, we discover that the sequence  $\{\Gamma^{\text{pseudo}}_i - \widetilde{\Gamma}_i \}^{n}_{i=1}$ forms a \textit{supermartingale} difference sequence. We justify this observation in the following lemma.

 \begin{lemma}
 \label{super_mar}
    For $i=1,...,n$, we define the difference $\Delta_i =    \Gamma^{\text{pseudo}}_i - \widetilde{\Gamma}_i$
    and the filtration $\mathcal{F}_{i} := (s_{\leq i},a_{\leq i},r_{\leq i-1},s^{\prime}_{\leq i-1})$, then it follows that 
    \$
\mathbb{E}[\Delta_i|\mathcal{F}_i] \leq 0,
    \$
where the strict equality holds if and only if the transition of the MDP, i.e., $\mathds{P}$, is deterministic. 
 \end{lemma}
 
Lemma \ref{super_mar} offers new evidence that the deviation of $\widetilde{\Gamma} - \Gamma^{\text{pseudo}}$ is quantifiable. Although the martingale structure does not hold, we can measure the deviation using \textit{supermartingale} concentration theory \citep{fan2015exponential}, and then quantify the statistical uncertainty in the \textit{pessimistic} CI estimation based on the relationship depicted in \eqref{transfer}.

\begin{thm}
\label{reg_sigma_find}
For any target policy $\pi$, Theorem \ref{reg_ci_thm} holds if we choose the lower confidence uncertainty deviation 
\#
\sigma^{L}_n = &\frac{C^{\star}\bar{V} + \lambda^{-} \mathbb{G}(C^{\star})}{1-\gamma}\sqrt{\frac{2\ln(4/\delta)}{n}}
\label{reg_final},
\#
where $C^{\star} = [(\mathbb{G}^{\prime})^{-1}(\min\{\bar{V}/\lambda^{-}, c_2 \})]^{+}$.
\end{thm}

In Theorem \ref{reg_sigma_find}, even without assuming any independence or weak dependence condition, our sample complexity is of order $\mathcal{O}(1/\sqrt{n})$, which aligns with the minimax rate in i.i.d. cases as discussed in \cite{duan2020minimax}. A closer examination of \eqref{reg_final} reveals that the uncertainty deviation $\sigma^{L}_n$ is fully independent of the complexity of $\Omega$. It is worth noting that equation \eqref{transfer} is valid for any function class $\Omega$. Consequently, it is not necessary to sacrifice the expressivity of the function class, for example, by restricting it to RKHS, in order to achieve point-wise uncertainty quantification. Following this principle, one can use a sufficiently expressive function class to minimize the model-misspecification error while maintaining efficient uncertainty quantification. The only cost is just the additional bias and uncertainty introduced by the discriminator function, for which it has been shown that this extra bias can be well-controlled by $\lambda^{-}$. We also note that the upper uncertainty $\sigma^{U}_n$ can be determined following Corollary $4.1$, as the confidence upper bound of the \textit{pessimistic} CI is almost identical to the one in the \textit{neutral} CI.

\begin{rmk}
We note that incorporating the discriminator function facilitates numerical optimization. The maximization over $\tau$ can be efficiently solved even though $\Omega$ is not a convex set. For instance, when $\Omega$ is a non-convex set but consists of monotonic functions or a linear transformation of strongly convex functions, the optimization can be solved easily.
\end{rmk}

\section{Optimization Algorithm}
\label{algo_sec}

In this section, we propose a stochastic approximation-type algorithm to solve the CI estimation problem. For simplicity of exposition, we discuss the optimization of the lower bound in the \textit{pessimistic} CI given by \eqref{reg_ci}. The upper confidence bound in \eqref{reg_ci} and the \textit{neutral} CI can be considered as a mirror or a special case of the proposed algorithm, respectively.

In function approximation settings, the $\Omega$ and $\mathcal{Q}$ are often represented by compact parametric functions in practice, either in linear or non-linear function classes \cite{sutton2018reinforcement}. We denote these parameters as $\psi$ and $\theta$ corresponding to $\Omega$ and $\mathcal{Q}$, respectively. One can express the parametric objective function for optimization as:
\#
\mathcal{L}(q_{\theta},\tau_{\psi}) = \frac{\mathbb{E}_{\mu}\left[\tau_{\psi}(s,a)(r(s,a)+q_{\theta}(s, a)-\gamma q_{\theta}(s^{\prime}, \pi))\right]}{1-\gamma} + q_{\theta}(s^0,\pi) - \frac{\lambda^{-}\mathbb{E}_{\mu}\left[\mathbb{G}(\tau_{\psi}(s,a))\right] }{1-\gamma}, 
\label{obj_func_prox}
\#
where $\mathbb{E}_{\mu}[\cdot]$ can be estimated by the observed data $\mathcal{D}_{i:n}$. Therefore, the optimization problem we aim to solve is $
\max_{\psi} \min_{\theta}  \mathcal{L}(q_{\theta},\tau_{\psi})$, 
which forms a saddle-point formulation, and we denote the saddle point as $(\psi^{*},\theta^{*})$. We observe that the inner minimization problem is relatively easy to solve. On one hand, when $\mathcal{Q}_{\theta}$ is within a finite ball of RKHS, the optimization yields a simple closed-form solution according to Theorem \eqref{rkhs_concen}. On the other hand, as demonstrated in Theorem {\color{blue}B.1} in the Appendix, the feature mapping class is sufficient for modeling $\mathcal{Q}$. The feature mapping class simplifies the optimization, making it efficiently solvable by various algorithms as discussed in \cite{sriperumbudur2011universality}.

In contrast, the more challenging aspect is optimizing $\tau_{\psi}$. Due to its complex structure, it demands a sufficiently flexible non-linear function approximation class, e.g., deep neural networks, for optimization \citep{jiang2020minimax}. Unfortunately, concavity typically does not hold for non-linear function approximation classes, and thus the outer maximization of $
\max_{\psi} \min_{\theta}  \mathcal{L}(q_{\theta},\tau_{\psi})$ is also affected. As a result, we need to develop a more efficient and convergent algorithm. From this perspective, our problem can be seen as solving a non-concave maximization problem, conditional on the solved global optimizer $\bar{q}_{\theta} := \argmin_{\theta} \mathcal{L}(q_{\theta},\tau_{\psi})$. Under this framework, we first study the gradients of the objective function with respect to ${\psi}$. Define $\bar{\mathcal{L}}(\tau_{\psi}) = \mathcal{L}( \bar{q}_{\theta},\tau_{\psi})$, then the gradient of $\bar{\mathcal{L}}(\tau_{\psi})$ with respect to $\psi$ satisfies 
\#
\nabla_{\psi}\bar{\mathcal{L}}(\tau_{\psi}) = \frac{\mathbb{E}_{\mu}[(r(s,a)+q_{\theta}(s, a)-\gamma q_{\theta}(s^{\prime}, \pi))\nabla_{\psi}\tau_{\psi}(s,a)]}{1-\gamma} - \frac{\lambda\mathbb{E}_{\mu}[\mathbb{G}^{\prime}(\tau(s,a))\nabla_{\psi}\tau_{\psi}(s,a)]}{1-\gamma},
\label{gradient_theory}
\#
where the corresponding stochastic gradient, denoted as $\widetilde{\nabla}_{\psi}\bar{\mathcal{L}}(\tau_{\psi})$, can be computed from offline dataset $\mathcal{D}_{1:n}$.
With the gradients provided in \eqref{gradient_theory}, we propose a stochastic approximation algorithm to update $\tau_{\psi}$. At each iteration, we update $\tau_{\psi}$ by solving the proximal mapping \cite{parikh2014proximal}: $
\text{Proj}_{\psi}(\psi^{*}, \nabla; D_{Berg}):= \argmax_{\psi} \{ \langle \psi, \nabla \rangle - D_{Berg}(\psi^{*}, \psi)\}$, 
where $\psi^{*}$ can be viewed as the current update of the parameter, $D_{Berg}(\cdot, \cdot)$ denotes the Bregman divergence as discussed in \cite{reem2019re}, and $\nabla$ represents the scaled stochastic gradient of the parameter of interest. In practice, we may consider using the Euclidean distance to reduce the computational burden. The details of the proposed algorithm are summarized in Algorithm \ref{prox_map_algo}. In the next section, we will demonstrate that our stochastic approximation algorithm is convergent with a sublinear rate even under non-linear (non-concave) settings. 


\begin{algorithm}[H]
	\caption{Proximal mapping Algorithm for Off-policy CI Estimation}
	\label{prox_map_algo}
	\begin{algorithmic}[1]
 \STATE \textbf{Input} offline data $\mathcal{D}_{1:n} := \{(s_i,a_i,r_i,s^{\prime}_i)\}^n_{i=1}$ and the initial state $s^0$.
			\STATE \textbf{Initialize} parameters of interest $\psi^{0}$ and $\theta^{0}$, 
   initial learning rate $\eta^{0}$,   hyperparameter $\lambda$, max iteration $T$, Bergman divergence function $D_{Berg}$. 
			\STATE \textbf{For} $t=1$ to $t = T$:
   	\STATE \; \ Update $\theta^{t}$ by solving 
   $$ \min_{\theta} \frac{1}{n}\sum^{n}_{i=1}\frac{\tau_{\psi^{t-1}}(s_i,a_i)(\gamma q_{\theta}(s^{\prime}_i,\pi) - q_{\theta}(s_i,a_i))}{1-\gamma} + q_{\theta}(s^0,\pi) - \frac{\sum^{n}_{i=1}\lambda^{-}\mathbb{G}(\tau_{\psi^{t-1}}(s_i,a_i))}{n(1-\gamma)}.$$
    \STATE \; \ Decay the stepsize $\eta^{t}$ of the rate $\mathcal{O}(t^{-1/4})$. 
      \STATE \; \ Compute the stochastic gradient with respect to $\psi$ as $\widetilde{\nabla}_{\psi}{\mathcal{L}}(\tau_{\psi},q_{\theta^{t}})$.
       \STATE \; \ Update $\psi^{t}$ by solving the prox-mapping: $\psi^{t} = \text{Proj}_{\psi}(\psi^{t-1}, \eta_{t}\widetilde{\nabla}_{\psi}{\mathcal{L}}(\tau_{\psi},q_{\theta^{t}});D_{Berg})$.
       \STATE \textbf{End for}
       \STATE \textbf{Output} $\widehat{\tau} =  \tau_{\psi^{T}}$ and  $\widehat{q} = q_{\theta^{T}}$.
\end{algorithmic}
\end{algorithm}

\section{Theoretical Properties}
\label{theoretical_sec}

In this section, we study the theoretical properties of the proposed CI estimations. In Theorems \ref{len_noreg}-\ref{len_reg}, we investigate the finite sample performance of the proposed method. Our theory shows that the sample complexity of learning a near-tight CI is of order $\mathcal{O}(1/\sqrt{n})$. The results widely hold for function approximation classes with finite 
pseudo-dimension. To the best of our knowledge, Theorems \ref{len_noreg}-\ref{len_reg} are the first finite sample bounds in CI estimation without explicitly requiring any independent or weakly dependent conditions on the data collection process. In Theorems \ref{error_robust_noreg}-\ref{error_robust_reg}, we study the robustness of the proposed algorithm with respect to the optimization and model-misspecification error, which makes our algorithm practical  in real-world applications. Moreover, we take a big step towards solving the open problem of the algorithmic convergence of the non-linear function approximation in offline RL. In Theorem \ref{conv_thm}, our algorithm is convergent to a stationary point at a sublinear rate even in non-concave settings. The convergent property guarantees the use of non-linear function classes, e.g., deep neural networks (DNNs), without divergence concerns.  
\subsection{Finite Sample Bound}

In this section, we present the finite-sample bound of the CI estimation. 
Before stating our results, we first make some mild assumptions.

\begin{assump}
\label{boundness_tau}
For $\tau \in \Omega$, suppose $0\leq \tau(s,a) \leq C$ for any $s, a$ and $C > 0$. 
\end{assump}

\begin{assump}
\label{good_tau}
For any target policy $\pi$ and $\nu \geq 0$, there exists a $\tau^{*} \in \Omega$ satisfying that 
\$
& \Big|\mathbb{P}_{n}\left\{\tau^{*}(s_i,a_i)(q(s_i,a_i)-\gamma q(s^{\prime}_i,\pi) -r_i)\right\}  \\
& \qquad \qquad \qquad - \mathbb{E}_{(s,a)\sim d^{\pi},s^{\prime} \sim \mathds{P}(\cdot|s,a)}\left[q(s,a)-\gamma q(s^{\prime},\pi) -r(s,a)\right]\Big| = \mathcal{O}\left(\frac{c_0}{n^{\frac{1}{1+\nu}}}\right),
\$
for any $q \in \mathcal{Q}$, where $c_0$ is some finite and positive constant. 
\end{assump}


Assumption \ref{boundness_tau} is a standard condition on function boundedness. Assumption \ref{good_tau} characterizes that there exists a ``good'' $\tau^{*} \in \Omega$, not necessarily exactly equal to the density ratio $d^{\pi}/\mu$, where the $\tau^{*}\cdot\mu$ is able to approximate $d^{\pi}$ well. This assumption is a much weaker assumption than the standard correct specification model assumption \citep{liu2018breaking,feng2020accountable}. The correct specification model assumption requires that the true model must be captured in $\Omega$. In contrast, our assumption only requires that the discrepancy between the limiting distribution of $\tau^{*}\cdot\mu$ and the distribution $d^{\pi}$ cannot be distinguished by the functional $\mathbb{E}_{s^{\prime} \sim \mathds{P}(\cdot|s,a)}[q(s,a) - \gamma q(s^{\prime},\pi) -r(s,a)]$, which is aligned with the principle in  \citep{chen2022offline}. As a special case, the assumption holds if  the limiting distribution of $\tau^{*}\cdot\mu$ converges to the $d^{\pi}$. The factor $\nu$ depends on the smoothness of $\mathcal{Q}$ \citep{tsybakov2009introduction}, and $\nu = 1$  when $\mathcal{Q}$ is a bounded ball in RKHS. The factor $c_0$ usually depends on the pseudo-dimension and radius of the specified function class.  In the following, 
we denote the confidence upper and lower bound optimized in \eqref{noreg_ci} over $\tau \in \Omega$ as $J^{+}_{n}(\pi)$ and $J^{-}_{n}(\pi)$, respectively. We first characterize the sample-dependent bound for the \textit{neutral} CI in the following.

\begin{thm}
\label{len_noreg}
Under Assumptions \ref{boundness_tau} and  \ref{good_tau}, then for any target policy $\pi$, the length of the estimated \textit{neutral} CI  satisfies 
\$
J^{+}_{n}(\pi) - J^{-}_{n}(\pi) \leq &
\frac{2c_0}{(1-\gamma)n^{\frac{1}{1+\nu}}}
 + \frac{C\bar{V}\mathcal{E}_{n}(\delta,\bar{V},D_{\Omega}) }{(1-\gamma)},
\$
where 
\#
\mathcal{E}_{n}(\delta,\bar{V},D_{\Omega}) := \left(\sqrt{\frac{2 \ln\frac{2e\left(D_{\Omega}+1\right)(4 e \bar{V}^2\sqrt{n})^{D_{\Omega}}}{\delta}}{n}} + \frac{4\ln\frac{2e\left(D_{\Omega}+1\right)(4 e \bar{V}^2\sqrt{n})^{D_{\Omega}}}{\delta}}{3n} \right),
\label{complex_rate}
\#
and $D_{\Omega}$ is the pseudo-dimension of the class $\Omega$. 
\end{thm}

The proof of Theorem \ref{len_noreg} relies on the empirical Freedman's bounds with a careful analysis of the variance term. When $\nu \leq 1$, the sample complexity of the generalization bound is of order $\mathcal{O}(1/\sqrt{n})$. This achieves the minimax optimal rate in off-policy evaluation \citep{duan2020minimax}. The bound can be potentially improved for some special classes, e.g., RKHS, which can be independent of the function space complexity. In some feature mapping classes, the finite sample bound might be vanishing at a sublinear rate, i.e., $\mathcal{O}(1/n)$ if $\lambda_{\max}(\Sigma_n) \ll \text{diam}_{\psi}$ as discussed in Corollary \ref{linear_class_noreg}. In the following, we present the result for the \textit{pessimistic} CI estimation in that $J^{+}_{n}(\pi)$ and $J^{-}_{n}(\pi)$ are the optimized confidence bounds in \eqref{reg_ci} within the class $\Omega$.  



\begin{thm}
\label{len_reg}
Under Assumptions \ref{boundness_tau} and  \ref{good_tau}, for any target policy $\pi$ we set $\lambda^{-} =\mathcal{O}(1/\sqrt{n})$, then the estimated \textit{pessimistic} CI satisfies that 
\$
(J(\pi) - J^{-}_{n}(\pi))(1-\gamma) \leq &\mathcal{O}\bigg(\frac{c_0}{n^{\frac{1}{1+\nu}}} + \sqrt{\frac{c_1^2}{n}} + \bar{V} \sqrt{\frac{(C^{\star})^2\ln\frac{4}{\delta}}{n}} \bigg), \\
(J^{+}_{n}(\pi) - J(\pi))(1-\gamma) \leq &
\mathcal{O}\bigg(\frac{c_0}{n^{\frac{1}{1+\nu}}} +  \frac{C\bar{V}\mathcal{E}_{n}(\delta/2,\bar{V},D_{\Omega}) }{(1-\gamma)}\bigg),
\$
where $C^{\star} = [(\mathbb{G}^{\prime})^{-1}(c_2)]^{+}$ and $\mathcal{E}_{n}(\delta,\bar{V},D_{\Omega})$ is defined in \eqref{complex_rate}. 
\end{thm}

Theorem \ref{len_reg} shows that it is sufficient to learn a near-tight confidence upper or lower bound with $\mathcal{O}(\sqrt{n})$ size of samples. Note that the error between the confidence lower bound and the discounted return, i.e., $J(\pi) - J^{-}(\pi)$, can be independent of function class complexity. To the best of our knowledge, this is the minimum estimation error bound that can be achieved. 


\subsection{Robustness to Optimization and Model Misspecification Errors}

In this section, we consider the setting where $\Omega \times \mathcal{Q}$ may not contain the ``good'' $\tau^{*}$, defined in Assumption \ref{good_tau}, and true action-value function $q^{\pi}$. We measure the approximation errors for the function classes $\Omega_{\psi}$ and $\mathcal{Q}_{\theta}$ as follows: 
\#
\varepsilon_{\tau} = \min_{\psi}\|\tau_{\psi} - \tau^{*}\|_{\infty}, \quad \varepsilon_{q} = \min_{\theta}\|q_{\theta} - q^{\pi}\|_{\infty}. 
\label{approx_error}
\#
In addition to the function approximation errors, we allow the existence of optimization errors. Let $(\widetilde{q}^{opt},\widetilde{\tau}^{opt})$ be the optimizer for the sample version of the objective function $\widehat{\mathcal{L}}(q_{\theta},\tau_{\psi})$ in \eqref{obj_func_prox}, then we define the optimization errors as 
\#
\widehat{\mathcal{L}}(\widetilde{q}^{opt},\widetilde{\tau}^{opt}) - \max_{\psi}\widehat{\mathcal{L}}(\widetilde{q}^{opt},\tau_{\psi}) \geq &  -\varepsilon_{opt}(\tau), \notag \\
\max_{\psi}\min_{\theta}\widehat{\mathcal{L}}(q_{\theta},\tau_{\psi}) -  \max_{\psi}\widehat{\mathcal{L}}(\widetilde{q}^{opt},\tau_{\psi})  \geq & -\varepsilon_{opt}(q), 
\label{opt_error}
\#
The above equations indicate that the max-min point of $\max_{\psi}\min_{\theta}\widehat{\mathcal{L}}(q_{\theta},\tau_{\psi})$ is approximated by the solution $\widehat{\mathcal{L}}(\widetilde{q}^{opt},\widetilde{\tau}^{opt})$. Similarly, we relax the min-max point of $\linebreak$ $\min_{\psi}\max_{\theta}\widehat{\mathcal{L}}(q_{\theta},\tau_{\psi})$ such that they can be obtained approximately as well, and we use the same error upper bounds, i.e., 
$\varepsilon_{opt}(q)$ and $\varepsilon_{opt}(\tau)$,  to eliminate unnecessary notations and improve readability. Under the definitions of \eqref{approx_error} and \eqref{opt_error}, we allow the existence of the model misspecification error and algorithm optimization error. In this case, we call our algorithm the \textit{robust} algorithm. In the following, Theorem \ref{error_robust_noreg} 
justifies that the robust algorithm is capable of learning a \textit{valid} and \textit{near-tight} neutral CI in polynomial sample complexity.   

\begin{thm}
\label{error_robust_noreg}
Under Assumptions \ref{boundness_tau} and  \ref{good_tau}, then for any target policy $\pi$, the estimated \textit{neutral} CI returned by the \textit{robust} algorithm is valid, and its length satisfies 
\$
J^{+}_{n}(\pi)  -  J^{-}_{n}(\pi) \leq \frac{1}{1-\gamma}\left(
\frac{2c_0}{n^{\frac{1}{1+\nu}}}
 + C\bar{V}\mathcal{E}_{n}(\delta,\bar{V},D_{\Omega}) + \varepsilon_{opt}+\varepsilon_{mis}\right),
\$
where $\varepsilon_{opt} = \varepsilon_{opt}(q) +\varepsilon_{opt}(\tau)$,  $\varepsilon_{mis} = 2\bar{V}\varepsilon_{\tau}+4C\varepsilon_{q}$, and $\mathcal{E}_{n}(\delta,\bar{V},D_{\Omega})$ defined in \eqref{complex_rate}. 
\end{thm}

Theorem \ref{error_robust_noreg} is an error-robust version of Theorem \ref{len_noreg}. The optimization error and approximation error are quantified within the finite-sample bound. For the optimization error, according to Theorem \ref{conv_thm}, Algorithm \ref{prox_map_algo} has optimization error of order $\mathcal{O}(1/T)$, where $T$ is the max number of the optimization iterations. For the approximation error, it depends on the specific structure of $\tau^{*}$ and $q^{\pi}$. For example, the expressivity of the feature mapping class might achieve an arbitrarily small approximation error  $\varepsilon_{q}$ as implied by Theorem {\color{blue}B.1} in the appendix. 
Blessed by the convergence of our algorithm under non-linear approximation settings, one may use a DNN class to model $\varepsilon_{\tau}$. Recent existing works show that DNNs enjoy universal approximation power  
in function approximations \citep{lu2020universal}, i.e., $\varepsilon_{\tau} \approx 0 $. In the following, we establish an error-robust finite sample analysis on the empirical length of the \textit{pessimistic} CI. 

\begin{thm}
\label{error_robust_reg}
Under Assumptions \ref{boundness_tau} and  \ref{good_tau}, for any target policy $\pi$ and we set $\lambda^{-} = \mathcal{O}(1/\sqrt{n})$,  the estimated \textit{pessimistic} CI returned by the \textit{robust} algorithm is valid, and $(J^{+}_{n}(\pi) - J^{-}_{n}(\pi))(1-\gamma)$ is bounded above by 
\$
\mathcal{O}&\Bigg(\frac{c_0}{n^{\frac{1}{1+\nu}}} + \sqrt{\frac{c_1^2}{n}} + \bar{V}\bigg\{C\mathcal{E}_{n}(\delta/2,\bar{V},D_{\Omega}) + \sqrt{\frac{(C^{\star})^2\ln\frac{4}{\delta}}{n}}+ \frac{c_1\sqrt{\ln\frac{4}{\delta}}}{n\bar{V}}\bigg\}  + \varepsilon_{opt} + \varepsilon^{\mathbb{G}}_{mis}  \Bigg),
\$
where $\varepsilon_{opt} = \varepsilon_{opt}(q) +\varepsilon_{opt}(\tau)$ and $\varepsilon^{\mathbb{G}}_{mis} = (c_2/\sqrt{n} + \bar{V})\varepsilon_{\tau}+C\varepsilon_{q}$.
\end{thm}

In comparison to Theorem \ref{error_robust_noreg}, there exists an additional model-misspecification error propagated to the finite-sample bound by the introduced discriminator function of order $\mathcal{O}(c_2\varepsilon_{\tau}/\sqrt{n})$. However, the strong convexity of the discriminator $\mathbb{G}(\cdot)$ facilitates the optimization, and thus our algorithm can achieve 
$\varepsilon_{opt}(\tau) = \mathcal{O}(1/T^2)$ at a faster rate than the rate $\mathcal{O}(1/T)$, when the class $\Omega$ is a special non-convex set consisting of monotonic
functions or linear transformations of strongly convex functions. 
A more concrete analysis of the approximation and optimization errors depends on the specific form of the function approximators, which is beyond the scope of this paper.

\subsection{Convergence Analysis}

In this section, we investigate the convergence of Algorithm \ref{prox_map_algo} when the $\tau_{\psi}$ is approximated by non-linear functions, e.g., DNNs, so that the concavity no longer holds. 
 In Theorem \ref{conv_thm}, we show that our algorithm converges sublinearly to a stationary point when stepsize is
diminishing. We first make the following regularity assumptions. 

\begin{assump}
\label{lip_tau}
For any $\tau_{\psi} \in \Omega_{\psi}$, $\tau_{\psi}$ is differentiable (not necessarily convex or concave), bounded from below,
$
\| \nabla_{\psi}\tau_{\psi_{1}}(s,a) - \nabla_{\psi}\tau_{\psi_{2}}(s,a) \| \leq L_{0}\| \psi_{1} - \psi_{2}\|, \; \text{for any} \; s,a 
$,
where $L_{0} < \infty$ is some universal Lipschitz constant and $\|\cdot\|$ denotes the Euclidean norm. 
\end{assump}

Assumption \ref{lip_tau} imposes the first-order smoothness condition on the specified function class. It requires that the gradient of $\tau_{\psi}$ be $L_0$-Lipschitz continuous, which is a standard assumption in the optimization literature \citep{bernstein2018signsgd,ghadimi2013stochastic}.


\begin{assump}
\label{lip_q}
$
    | q_{\theta_1}(s,a) - q_{\theta_2}(s,a) | \leq L_{0}\| \theta_1 - \theta_2\|,  \; \text{for any} \; s,a, \; \text{and} \; q_{\theta} \in \mathcal{Q}_{\theta}$.
\end{assump}
Assumption \ref{lip_q} holds for a wide range of function approximation classes, including feature mapping space with smooth basis functions, non-linear approximation classes, DNNs with Leaky ReLU activation function, or spectral normalization on ReLU activation \citep{izmailov2018averaging}.

\begin{assump}
\label{bound_gradient}
The gradient of function $\tau_{\psi}(\cdot)$ evaluated at saddle point $\psi^{*}$ is bounded above; i.e., 
$\nabla_{\psi}\tau_{\psi^{*}}(s,a) < c_3$ uniformly over $(s,a)$ for some finite and positive constant $c_3$. 
\end{assump}

Assumption \ref{bound_gradient} is a much weaker assumption compared to the bounded variance of stochastic gradients assumption which is commonly made 
in the existing literature \citep{reddi2016stochastic,mokhtari2020unified}. Our relaxation is guaranteed by the careful analysis of the variance of the stochastic gradient, which is shown to be upper-bounded by the suboptimality of the current iteration and the stochastic gradient at the saddle point. In the following, we derive the convergence rate for Algorithm \ref{prox_map_algo}, which holds for non-concave function approximation class $\Omega_{\psi}$, e.g., DNNs.

\begin{thm}
\label{conv_thm}
Under Assumptions \ref{boundness_tau} and  \ref{lip_tau}-\ref{bound_gradient}, suppose Algorithm \ref{prox_map_algo} runs $T \geq1$ rounds with stepsize $\eta^{t} = \min\{(tT)^{-1/4}\sqrt{2\Lambda/\sigma^2_{\max}L_1},1/L_1\}$ for $t=1,...,T$ and Euclidean distance is used for Bergman divergence. If we pick up the solution output $\psi^{T^{\star}}$ following the probability mass function $P(T^{\star}=t) = \frac{2\eta^{t}-(\eta^{t})^2L_1}{\sum^{T}_{t=1}(2\eta^{t}-(\eta^{t})^2L_1)}$, then
\#
\mathbb{E}[\|\nabla_{\psi}\bar{\mathcal{L}}(\tau_{\psi^{T^{\star}}}) \|^2] \leq \frac{2\Lambda L_1}{T} + \frac{\sqrt{2\Lambda L_1\sigma^2_{\max}}}{T^{3/4}} + \sqrt{\frac{2\Lambda L_1\sigma^2_{\max}}{T}},
\label{conv_rate}
\#
where $\bar{\mathcal{L}}(\tau_{\psi})$ is defined in \eqref{gradient_theory} and $\Lambda:= \bar{\mathcal{L}}(\tau_{\psi^{0}}) -  \min_{\psi}\bar{\mathcal{L}}(\tau_{\psi})$ measures the distance of the initial and optimal solution, $L_1$ is Lipschitz constant depending on $c_2,c_3,M, L_0, \bar{V}$, and the variance of the stochastic gradient is bounded above by 
\$
\sigma_{\max} := \max_{t\in 1:T}\sqrt{ c_4\|\widetilde{\theta}(\psi^{t}) - \theta^{*}\|^2 + c_5\|\psi^{t} - \psi^{*}\|^2},
\$
for some constants $c_4$ depending on $c_3,L_0,\gamma$; and $c_5$ depending on $c_2,L_0,\bar{V},\lambda^{-}$ and $\gamma$. Here, $\widetilde{\theta}(\psi^{t})$ is the optimizer for $\mathcal{L}(q_\theta,\tau_{\psi^{t}})$.
\end{thm}

Theorem \ref{conv_thm} implies that Algorithm \ref{prox_map_algo} can converge sublinearly to a stationary point if the $\sigma_{\max}$ is sufficiently small. The rate of convergence is also affected by the smoothness of the class $\Omega_{\psi}$ and the distance of the initial and optimal solution. We possess a nearly optimal rate since the third term in \eqref{conv_rate} is unimprovable for nonconvex problems for gradient descent methods, as shown in \cite{kornowski2022oracle}. In the algorithm, we initially use decreasing stepsizes to enable early stopping. However, as the algorithm progresses and more local information about the gradient is obtained, increasing stepsizes, such as $\min\{1/L_1, \mathcal{O}(\sqrt{t}/T)\}$, may be more effective. This increasing stepsize strategy can also result in a convergence rate similar to the obtained one in \eqref{conv_rate}.

\section{Simulation Studies}
\label{simu_sec}

In this section, we evaluate the empirical performance of our methods using two synthetic datasets: CartPole benchmark environment from the
OpenAI Gym \citep{brockman2016openai} and the simulation environment \citep{liao2020off}. In Section \ref{tight_ana}, we study the tightness of the estimated CIs. In Section \ref{robust_shift}, we investigate the policy evaluation performance based on the estimated CIs in scenarios with large distributional shifts. Additionally, we conduct a sensitivity analysis with respect to the hyperparameter $\lambda$ in the \textit{pessimistic} CI. We compare the proposed CI with 
several state-of-the-art approaches including Boostrapping Fitted Q-evaluation (B-FQE;  \cite{hao2021bootstrapping}), Accountable Bellman Evaluation (ABE; \cite{feng2019kernel}) and the stepwise importance sampling-based
estimator, HCOPE proposed by \cite{thomas2015high}.  The true discounted return of the target policy $\pi$ is computed by Monte Carlo approximations \citep{luckett2020estimating}. For the simulation environment setting, 
the system dynamics are given by
$$
\begin{aligned}
s^{t+1} & =\left(\begin{array}{cc}
0.75\left(2 a^{ t}-1\right) & 0 \\
0 & 0.75\left(1-2 a^{ t}\right)
\end{array}\right) s^{t}+\left(\begin{array}{cc}
0 & 1 \\
1 & 0
\end{array}\right)\odot
s^{t}{s^{t}}^{\top}\mathbb{I}_{2\times 1} + 
\varepsilon^t, \\
r^{t} & ={s^{t+1}}^{\top}\left(\begin{array}{l}
2 \\
1
\end{array}\right)-\frac{1}{4}\left(2 a^{ t}-1\right) + ({s^{t+1}}^{\top}s^{t+1})^{\frac{3}{2}}\odot\left(\begin{array}{c}
0.25 \\
0.5 
\end{array}\right),
\end{aligned}
$$
for $t \geq 0$, where $\odot$ denotes the Hadamard product, $\mathbb{I}$ is the identity matrix, the noise $\left\{\varepsilon^t\right\}_{t \geq 0} \stackrel{i i d}{\sim} N\left({0}_{2\times 1}, 0.25\mathbb{I}_{2\times 2}\right)$ and the initial state variable $s^{0} \sim N\left({0}_{2\times 1}, 0.25\mathbb{I}_{2\times 2}\right)$. The transition dynamic mainly follows the design in \cite{liao2020off}, but the reward function we consider here is more complex. For the basic setup of CartPole environment, we refer the readers to \cite{brockman2016openai} for more details. We fix the confidence
level at $\delta=0.05$, and the discount factor $\gamma = 0.95$ to construct CIs. In particular, for implementing \text{pessimistic} CI estimation, we set the discriminator function as a quadratic form, i.e., $\mathbb{G}(x) = \frac{1}{2}(x-1)^2$.

\subsection{Empirical Tightness Analysis of Estimated Interval}
\label{tight_ana}

In this section, we calculate the width of the estimated CIs. We 
consider different sample size settings. Note that the setting $n=200$ is an extremely small sample size in offline RL problems \citep{levine2020offline}. 

In the CartPole environment, we follow \cite{uehara2020minimax} to learn a near-optimal policy using a softmax policy class
\citep{haarnoja2018softac} with a temperature parameter $\alpha=1$ as the behavior policy. In addition, we set $\alpha=0.4$ and $\alpha=0.2$ for constructing two target policies. We note that the target policies are more deterministic than the behavior policies, and a smaller $\alpha$ implies a larger difference between the behavior policy and the target policy. In the simulation environment, we consider a completely randomized study and set $\{a^{t}\}_{t\geq0}$ to be i.i.d Bernoulli random variables with expectation $0.5$. The target policy we consider is designed as $\pi(a=1| s) =p$ if $s_{[1]} + s_{[2]} > 0$ and $1-p$ otherwise, where $s_{[j]}$ denotes the $j$-th element of the vector $s$. The two target policies are generated by $p=0.9$ and $0.7$, respectively. In both environments, we use a feedforward neural network of $2$ hidden
layers with $256$ units for each layer to approximate $\Omega$ in the \textit{pessimistic} CI estimation, and an RKHS finite representer for function approximation in the \textit{neutral} CI estimation.



\begin{figure}[!htb]
   \begin{minipage}{0.6\textwidth}
     \centering
     \includegraphics[width=.9\linewidth]{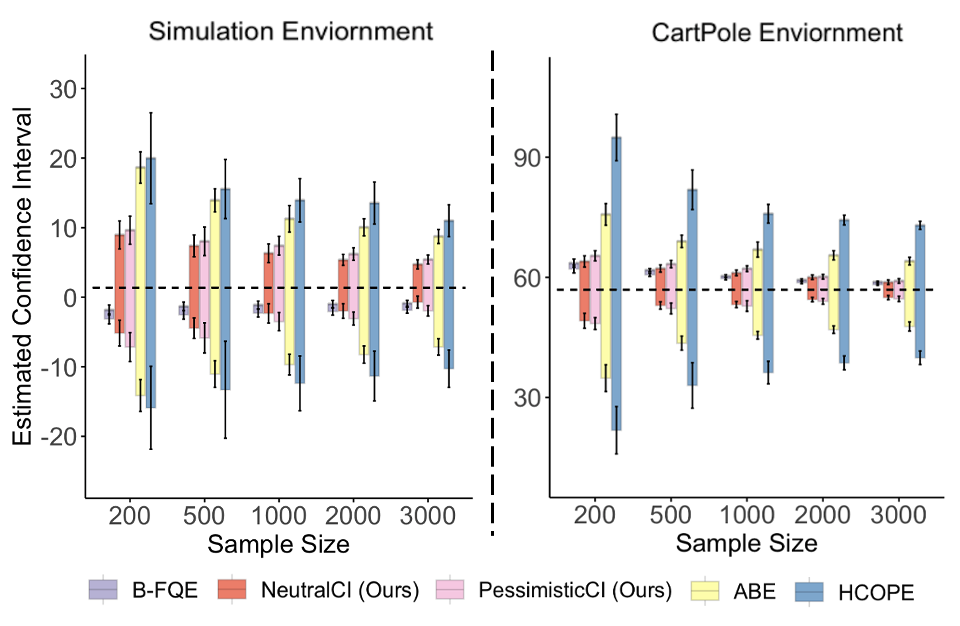}
\vspace{-3.5mm}
     \caption{The estimated CIs for the target policy $1$ of the simulation and CartPole environments over different methods. The experiments are repeated $50$ times with random seeds. The horizontal dashed line represents the Monto-Carlo true return for the target policy. The threshold parameter $\lambda$ for \textit{pessimistic} CI is fixed to be $1$.
     }\label{cartpole}
   \end{minipage}\hfill
   \begin{minipage}{0.37\textwidth}
     \centering
     \includegraphics[width=0.98\linewidth]{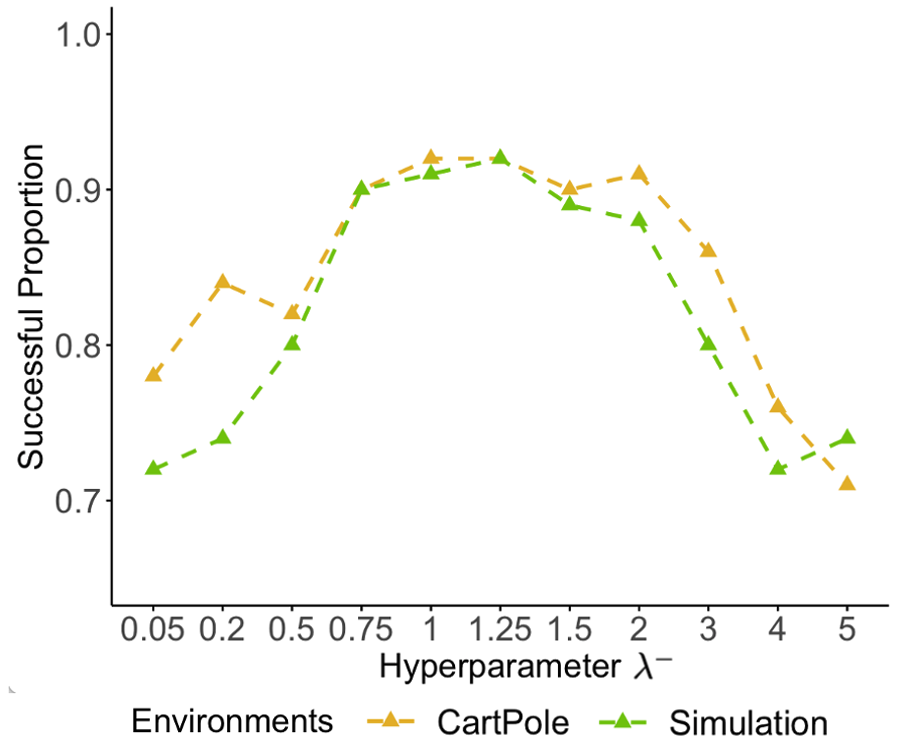}
     \caption{The sensitivity analysis of the hyperparameter $\lambda^{-}$ on the successful ranking proportion in the two synthetic environments with $50$ repeated experiments. }\label{sense}
   \end{minipage}
\end{figure}

The results of the target policy $1$ with respect to the simulation
 and CartPole  environment is depicted in Figures \ref{cartpole}, where the results for target policy $2$ are provided in the appendix. We summarize our findings as follows. First, our proposed \textit{neutral} and \textit{pessimistic} CIs yield much tighter bounds than ABE and HCOPE, especially in the small sample size setting (e.g., $n=200,500$). Although B-FQE achieves tighter intervals than ours, it is highly biased and fails to capture the Monte Carlo true reward of the target policy. The tightness of our CIs is mainly due to the fact that our framework breaks the curse of the bias and uncertainty tradeoff. This allows us to choose an expressive neural network or finite RKHS function class to model $\Omega$ and reduce model-misspecification bias without worrying about an increase in uncertainty deviation.  
Another reason for the performance gain may be that our proposed method does not assume any i.i.d. or weakly dependent data assumptions, which is not the case for the competing methods.  When applied directly to interdependent data, the competing methods often lead to larger bias and an increased interval width \citep{duchi2021statistics}. Lastly, our proposed method inherently quantifies the model-misspecification error in the CI and is thus robust to function approximation error. This distinguishes our method from the competing methods, which fail to quantify the model-misspecification error and suffer from large bias when the true model is not correctly specified. 

Regarding the comparison between the \textit{neutral} and \textit{pessimistic} CIs, we observe that the \textit{pessimistic} CI is generally wider than the \textit{neutral} one in the simulation environment. This might be because modeling $\Omega$ in RKHS has already achieved sufficient  expressivity to capture the true model in the smooth MDP. Therefore, the \textit{pessimistic} CI indeed does not have any additional    advantages in minimizing model-misspecification errors instead of inheriting additional bias from the discriminator function. In contrast, the smoothness MDP condition does not hold in the CartPole environment, and the \textit{neutral} CI incorporates more bias. This leads to the \textit{neutral} CI sharing almost the same tightness as the estimated \textit{pessimistic} CI.


\subsection{Robustness to Distributional Shifts}
\label{robust_shift}

In this section, we aim to evaluate the performance of the proposed CIs in handling distributional shifts. In the context of offline reinforcement learning, a common safe optimization criterion is to maximize the worst-case performance among a set of statistically plausible models \citep{kumar2019stabilizing,jin2021pessimism}. This pessimism principle is closely related to robust MDPs which can handle distributional shift \citep{xie2021bellman,jin2021pessimism}. Our proposed framework naturally provides a lower bound to implement the pessimism principle. To evaluate the proposed method, we conduct experiments with a set of target policies that have varying degrees of distributional shift. We intentionally design these target policies to have almost the same Monte Carlo true rewards. Therefore, users will select the policies close to the behavior policy, i.e., those with a small distributional shift. We use the estimated confidence lower bound to rank the target policies, and the criterion for good performance is to correctly rank the policies according to their degree of distributional shift.

In the experiment design of robustness to distributional shift, we still use the soft actor-critic algorithm \citep{haarnoja2018softac} to construct the softmax policies. In contrast to the constructions in the last section, here we learn a near-optimal policy, with a very small temperature parameter $\alpha=0.05$, as the behavior policy in the CartPole environment. This indicates that the behavior policy is less explorable and has a higher probability of causing large distributional shifts in the policy evaluation step. To evaluate the proposed method, we generate three different near-optimal policies as the target policies, denoted as $\pi^{\text{1}},\pi^{\text{2}},\pi^{\text{3}}$. The temperature parameter used for learning the three policies increases from $\alpha=0.2$ for $\pi^{\text{1}}$ to $\alpha=0.5$ for $\pi^{\text{2}}$ and then $\alpha=0.8$ for $\pi^{\text{3}}$, indicating that $\pi^{\text{1}}$ has the smallest distributional shift while $\pi^{\text{3}}$ has the largest. We carefully control the true rewards of the three policies within a very small gap of 1, so the correct ranking of the three target policies should follow $\pi^{\text{1}}$, then $\pi^{\text{2}}$, and $\pi^{\text{3}}$. In the simulation environment, we apply the same principle for experiment design. Specifically, we generate the target policy $\pi^{\text{1}}$ by adding a small random noise to the behavior policy, resulting in the smallest distributional shift. The remaining two target policies are designed to be more explorable than $\pi^{\text{1}}$.

\begin{table}[H]
\vspace{-0.5cm}
\centering
{\footnotesize
\begin{tabular}{c|cccccc}
\hline
& NeutralCI & PessmisticCI & B-FQE & ABE & HCOPE \\ \hline
Simulation & 71/100 & 92/100 & 64/100& 53/100 & 37/100 \\ \hline
Cartpole & 78/100 & 95/100 & 71/100  & 67/100  & 38/100 \\ 
\hline
\end{tabular} 
}
 \caption{The proportion of the successful policy ranking by the estimated confidence lower bound over different methods in $100$ repeated experiments. For each experiment, success count is only made when the absolute ranking for the three target policies is correct, otherwise, the count is a failure.}
  \label{rank_percentage}
  \vspace{-0.5cm}
\end{table}

 The results in Table \ref{rank_percentage} demonstrate that the \textit{pessimistic} CI achieves the best performance in comparison to the other approaches. The discriminator in the \textit{pessimistic} approach helps to mitigate the bias estimation issue caused by distributional shift. 
Among the remaining approaches, we find the result of the proposed \textit{neutral} CI is consistent with B-FQE. The decent performance of the \textit{neutral} CI is 
due to the relatively accurate estimation of the confidence lower bound, while the performance of B-FQE mainly relies on the distributional consistency guarantee which provides certain robustness to distributional shift \citep{hao2021bootstrapping}.



We also provide a sensitivity analysis of the hyperparameter $\lambda^{-}$ in Figure \ref{sense}. The $\lambda^{-}$'s values vary in a wide range $[0.05,5]$, and the successful percentages of the ranking tasks are reported accordingly. In general, the successful ranking percentage is not very sensitive to the choice of $\lambda^{-}$. It can be seen that the desired results mainly concentrate in the range $[0.75,2]$. Too small $\lambda^{-}$ leads to insufficient distributional shift adjustment, but too large $\lambda^{-}$ results in overly pessimistic reasoning.

\section{OhioT1DM Case Study}
\label{real_sec}
We apply the proposed methods  to the 
Ohio type 1 diabetes (OhioT1DM) mobile health study \citep{marling2020ohiot1dm}, which contains six patients with type 1 diabetes, each patient with eight weeks of life-event data including health status measurements and insulin injection dosage. As each individual patient has distinctive glucose dynamics, we follow \cite{zhu2020causal} in regarding each patient's data as an independent dataset where the data from each day is a single trajectory. We summarize the collected measurements over $60$-min intervals such that the maximum length of the horizon is $24$ and the total number of the transition pairs in each patient's dataset is around $n=360$ after removing missing samples and outliers. The state variable $s^{t}$ is set to be a three-dimensional vector including the average blood glucose levels $s^{t}_{[1]}$, the average heart rate $s^{t}_{[2]}$ and the total carbohydrates $s^{t}_{[3]}$ intake during the period time $[t-1,t]$. Here, the reward is defined as the average of the index of glycemic control \citep{rodbard2009interpretation} between time $t-1$ and $t$, measuring the health status of the patient's glucose level. That is
$
{r^{t}}=-\frac{\mathbb{I}({s^{t}_{[1]}}>140)|{s^{t}_{[1]}}-140|^{1.10} + \mathbb{I}({s^{t}_{[1]}}<80)({s^{t}_{[1]}} - 80)^{2}}{30},
$
which implies that reward $r^{t}$ is non-positive and a larger value is preferred. We discretize the insulin dose level to $5$ grids, forming the action space $\mathcal{A} = \{0,1,...,4\}$, and we set $\gamma=0.9, \delta=0.05$ for estimating the CIs.

\begin{figure}[h]
\vspace{-0.5cm}
\centering
\includegraphics[scale=0.24]{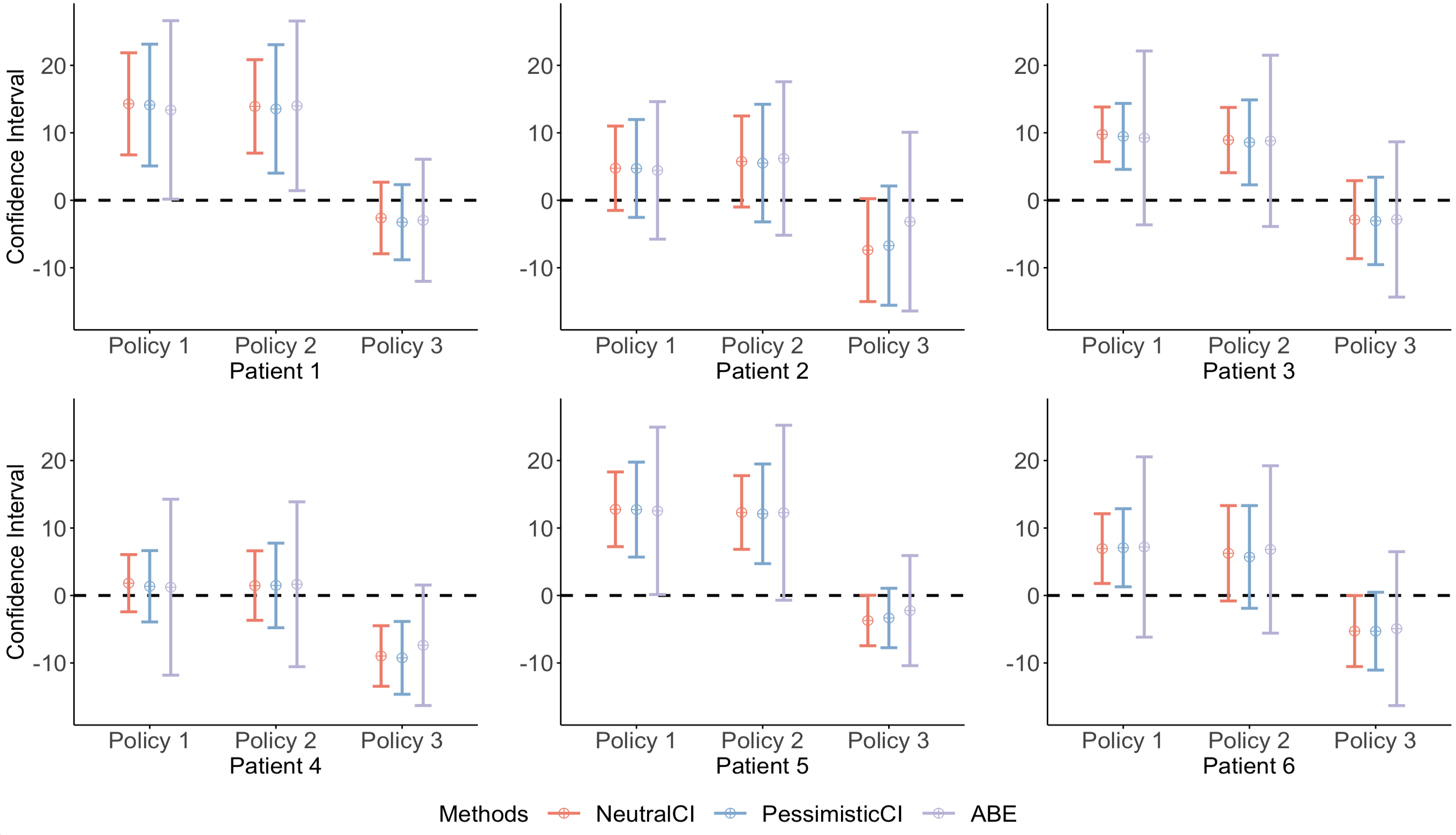}
\caption {The estimated CI for the difference between the three target policies and the behavior policy over $6$ patients in OhioT1DM dataset. 
policy 1 is learned by
by a distributional shift robust policy optimization algorithm, conservative q-learning  \citep{kumar2020conservative}, policy 2 is learned by the soft actor-critic algorithm \citep{haarnoja2018softac}, and policy 3 is set to be a randomized policy.}
\label{real_data}
\vspace{-0.5cm}
\end{figure}

We consider three different target policies for evaluation, where the design settings are described in Figure \ref{real_data}. The true value of $\pi^{\text{3}}$ shall be lower than the other two policies and also lower than the behavior policy made by the clinicians. For the policies $\pi^{\text{1}}$ and $\pi^{\text{2}}$, they are both near-optimal and should achieve similar performance regarding the mean value \citep{kumar2020conservative}. However, as the conservative q-learning algorithm is risk-sensitive to distributional shift, one can expect policy $\pi^{\text{1}}$ to have a higher lower bound in the estimated CI.

In Figure \ref{real_data}, we plot the estimated CIs by our method and the ABE method. We summarize our findings as
follows. In terms of the validity of the proposed CI estimation, both the proposed \textit{neutral} and \textit{pessimistic} CI are overlapped with, and included, in the CI estimated by ABE. In addition, our CIs for the policies $\pi^{\text{1}}$ and $\pi^{\text{2}}$ are above $0$ and the estimated intervals for $\pi^{\text{3}}$ are below $0$. This is consistent with the fact that policy $\pi^{\text{1}}$ and $\pi^{\text{2}}$ are designed as near-optimal policies which should be better than the behavior policy. In contrast, policy $\pi^{3}$ is a randomized policy and thus will be worse than the policy induced by the clinicians. The above findings confirm that our estimated CIs are valid. In terms of tightness, our CIs are narrower than the ones estimated by ABE. This confirms that the proposed method solves the issue of the tradeoff between bias and uncertainty. In addition, ABE requires i.i.d sampled offline data, thus failing with the complicated real environment and leading to wide estimated intervals. In terms of performance against distributional shifts among all methods, the estimated \textit{pessimistic} CI identifies a consistently higher confidence lower bound for the distributional shift robust policy $\pi^{\text{1}}$ over all $6$ patients. This implies that the \textit{pessimistic} CI can efficiently evaluate the target policy by taking the distributional shift information into account. 
In conclusion, our analysis suggests  applying the proposed method could potentially improve some patients’ health status efficiently and reliably, especially in environments with large distributional shifts.  

\section{Discussion}
\label{dis_sec}

This paper investigates the off-policy confidence interval estimation problem in behavior-agnostic settings with the minimum data collection assumption. The proposed framework incorporates distributional shift information, which makes the estimated confidence interval robust in learning from pre-collected offline datasets. We propose a unified error analysis and point out the issue regarding the tradeoff between evaluation bias and statistical uncertainty. Then a tradeoff estimation approach is developed to solve this issue, which leads to possibly the tightest confidence interval estimations. Several improvements and extensions are worth exploring in the future. First, we may extend the proposed framework to the environment with an unobservable confounder. In addition, our numerical studies have demonstrated the potential of the proposed method in ranking tentative target policies, which could further enhance designing safe policy improvement or policy optimization algorithms based on the current framework. This could be a potentially interesting direction to explore in the future.

\bibliographystyle{imsart-number} 
\bibliography{mycite}       

\newcommand{\noop}[1]{}
\begin{thebibliography}{75}

\bibitem{agarwal2019reinforcement}
\begin{barticle}[author]
\bauthor{\bsnm{Agarwal},~\bfnm{Alekh}\binits{A.}}, \bauthor{\bsnm{Jiang},~\bfnm{Nan}\binits{N.}}, \bauthor{\bsnm{Kakade},~\bfnm{Sham~M}\binits{S.~M.}} \AND \bauthor{\bsnm{Sun},~\bfnm{Wen}\binits{W.}}
(\byear{2019}).
\btitle{Reinforcement learning: Theory and algorithms}.
\bjournal{CS Dept., UW Seattle, Seattle, WA, USA, Tech. Rep}
\bpages{4--10}.
\end{barticle}
\endbibitem

\bibitem{bach2017breaking}
\begin{barticle}[author]
\bauthor{\bsnm{Bach},~\bfnm{Francis}\binits{F.}}
(\byear{2017}).
\btitle{Breaking the curse of dimensionality with convex neural networks}.
\bjournal{The Journal of Machine Learning Research}
\bvolume{18}
\bpages{629--681}.
\end{barticle}
\endbibitem

\bibitem{baird1995residual}
\begin{bincollection}[author]
\bauthor{\bsnm{Baird},~\bfnm{Leemon}\binits{L.}}
(\byear{1995}).
\btitle{Residual algorithms: Reinforcement learning with function approximation}.
In \bbooktitle{Machine Learning Proceedings 1995}
\bpages{30--37}.
\bpublisher{Elsevier}.
\end{bincollection}
\endbibitem

\bibitem{bernstein2018signsgd}
\begin{binproceedings}[author]
\bauthor{\bsnm{Bernstein},~\bfnm{Jeremy}\binits{J.}}, \bauthor{\bsnm{Wang},~\bfnm{Yu-Xiang}\binits{Y.-X.}}, \bauthor{\bsnm{Azizzadenesheli},~\bfnm{Kamyar}\binits{K.}} \AND \bauthor{\bsnm{Anandkumar},~\bfnm{Animashree}\binits{A.}}
(\byear{2018}).
\btitle{signSGD: Compressed optimisation for non-convex problems}.
In \bbooktitle{International Conference on Machine Learning}
\bpages{560--569}.
\bpublisher{PMLR}.
\end{binproceedings}
\endbibitem

\bibitem{brockman2016openai}
\begin{barticle}[author]
\bauthor{\bsnm{Brockman},~\bfnm{Greg}\binits{G.}}, \bauthor{\bsnm{Cheung},~\bfnm{Vicki}\binits{V.}}, \bauthor{\bsnm{Pettersson},~\bfnm{Ludwig}\binits{L.}}, \bauthor{\bsnm{Schneider},~\bfnm{Jonas}\binits{J.}}, \bauthor{\bsnm{Schulman},~\bfnm{John}\binits{J.}}, \bauthor{\bsnm{Tang},~\bfnm{Jie}\binits{J.}} \AND \bauthor{\bsnm{Zaremba},~\bfnm{Wojciech}\binits{W.}}
(\byear{2016}).
\btitle{OpenAI gym}.
\bjournal{arXiv preprint arXiv:1606.01540}.
\end{barticle}
\endbibitem

\bibitem{challen2019artificial}
\begin{barticle}[author]
\bauthor{\bsnm{Challen},~\bfnm{Robert}\binits{R.}}, \bauthor{\bsnm{Denny},~\bfnm{Joshua}\binits{J.}}, \bauthor{\bsnm{Pitt},~\bfnm{Martin}\binits{M.}}, \bauthor{\bsnm{Gompels},~\bfnm{Luke}\binits{L.}}, \bauthor{\bsnm{Edwards},~\bfnm{Tom}\binits{T.}} \AND \bauthor{\bsnm{Tsaneva-Atanasova},~\bfnm{Krasimira}\binits{K.}}
(\byear{2019}).
\btitle{Artificial intelligence, bias and clinical safety}.
\bjournal{BMJ Quality \& Safety}
\bvolume{28}
\bpages{231--237}.
\end{barticle}
\endbibitem

\bibitem{chen2022offline}
\begin{binproceedings}[author]
\bauthor{\bsnm{Chen},~\bfnm{Jinglin}\binits{J.}} \AND \bauthor{\bsnm{Jiang},~\bfnm{Nan}\binits{N.}}
(\byear{2022}).
\btitle{Offline reinforcement learning under value and density-ratio realizability: the power of gaps}.
In \bbooktitle{Uncertainty in Artificial Intelligence}
\bpages{378--388}.
\bpublisher{PMLR}.
\end{binproceedings}
\endbibitem

\bibitem{chen2022well}
\begin{binproceedings}[author]
\bauthor{\bsnm{Chen},~\bfnm{Xiaohong}\binits{X.}} \AND \bauthor{\bsnm{Qi},~\bfnm{Zhengling}\binits{Z.}}
(\byear{2022}).
\btitle{On well-posedness and minimax optimal rates of nonparametric q-function estimation in off-policy evaluation}.
In \bbooktitle{International Conference on Machine Learning}
\bpages{3558--3582}.
\bpublisher{PMLR}.
\end{binproceedings}
\endbibitem

\bibitem{choi2003feature}
\begin{barticle}[author]
\bauthor{\bsnm{Choi},~\bfnm{Euisun}\binits{E.}} \AND \bauthor{\bsnm{Lee},~\bfnm{Chulhee}\binits{C.}}
(\byear{2003}).
\btitle{Feature extraction based on Bhattacharyya distance}.
\bjournal{Pattern Recognition}
\bvolume{36}
\bpages{1703--1709}.
\end{barticle}
\endbibitem

\bibitem{dai2020coindice}
\begin{barticle}[author]
\bauthor{\bsnm{Dai},~\bfnm{Bo}\binits{B.}}, \bauthor{\bsnm{Nachum},~\bfnm{Ofir}\binits{O.}}, \bauthor{\bsnm{Chow},~\bfnm{Yinlam}\binits{Y.}}, \bauthor{\bsnm{Li},~\bfnm{Lihong}\binits{L.}}, \bauthor{\bsnm{Szepesv{\'a}ri},~\bfnm{Csaba}\binits{C.}} \AND \bauthor{\bsnm{Schuurmans},~\bfnm{Dale}\binits{D.}}
(\byear{2020}).
\btitle{CoinDICE: Off-policy confidence interval estimation}.
\bjournal{Advances in Neural Information Processing Systems}
\bvolume{33}
\bpages{9398--9411}.
\end{barticle}
\endbibitem

\bibitem{dai2017boosting}
\begin{barticle}[author]
\bauthor{\bsnm{Dai},~\bfnm{Bo}\binits{B.}}, \bauthor{\bsnm{Shaw},~\bfnm{Albert}\binits{A.}}, \bauthor{\bsnm{He},~\bfnm{Niao}\binits{N.}}, \bauthor{\bsnm{Li},~\bfnm{Lihong}\binits{L.}} \AND \bauthor{\bsnm{Song},~\bfnm{Le}\binits{L.}}
(\byear{2017}).
\btitle{Boosting the actor with dual critic}.
\bjournal{arXiv preprint arXiv:1712.10282}.
\end{barticle}
\endbibitem

\bibitem{duan2020minimax}
\begin{binproceedings}[author]
\bauthor{\bsnm{Duan},~\bfnm{Yaqi}\binits{Y.}}, \bauthor{\bsnm{Jia},~\bfnm{Zeyu}\binits{Z.}} \AND \bauthor{\bsnm{Wang},~\bfnm{Mengdi}\binits{M.}}
(\byear{2020}).
\btitle{Minimax-optimal off-policy evaluation with linear function approximation}.
In \bbooktitle{International Conference on Machine Learning}
\bpages{2701--2709}.
\bpublisher{PMLR}.
\end{binproceedings}
\endbibitem

\bibitem{duchi2021statistics}
\begin{barticle}[author]
\bauthor{\bsnm{Duchi},~\bfnm{John~C}\binits{J.~C.}}, \bauthor{\bsnm{Glynn},~\bfnm{Peter~W}\binits{P.~W.}} \AND \bauthor{\bsnm{Namkoong},~\bfnm{Hongseok}\binits{H.}}
(\byear{2021}).
\btitle{Statistics of robust optimization: A generalized empirical likelihood approach}.
\bjournal{Mathematics of Operations Research}
\bvolume{46}
\bpages{946--969}.
\end{barticle}
\endbibitem

\bibitem{fan2015exponential}
\begin{barticle}[author]
\bauthor{\bsnm{Fan},~\bfnm{Xiequan}\binits{X.}}, \bauthor{\bsnm{Grama},~\bfnm{Ion}\binits{I.}} \AND \bauthor{\bsnm{Liu},~\bfnm{Quansheng}\binits{Q.}}
(\byear{2015}).
\btitle{Exponential inequalities for martingales with applications}.
\bjournal{Electronic Journal of Probability}
\bvolume{20}
\bpages{1--22}.
\end{barticle}
\endbibitem

\bibitem{feng2019kernel}
\begin{barticle}[author]
\bauthor{\bsnm{Feng},~\bfnm{Yihao}\binits{Y.}}, \bauthor{\bsnm{Li},~\bfnm{Lihong}\binits{L.}} \AND \bauthor{\bsnm{Liu},~\bfnm{Qiang}\binits{Q.}}
(\byear{2019}).
\btitle{A kernel loss for solving the Bellman equation}.
\bjournal{arXiv preprint arXiv:1905.10506}.
\end{barticle}
\endbibitem

\bibitem{feng2020accountable}
\begin{binproceedings}[author]
\bauthor{\bsnm{Feng},~\bfnm{Yihao}\binits{Y.}}, \bauthor{\bsnm{Ren},~\bfnm{Tongzheng}\binits{T.}}, \bauthor{\bsnm{Tang},~\bfnm{Ziyang}\binits{Z.}} \AND \bauthor{\bsnm{Liu},~\bfnm{Qiang}\binits{Q.}}
(\byear{2020}).
\btitle{Accountable off-policy evaluation with kernel bellman statistics}.
In \bbooktitle{International Conference on Machine Learning}
\bpages{3102--3111}.
\bpublisher{PMLR}.
\end{binproceedings}
\endbibitem

\bibitem{feng2021non}
\begin{barticle}[author]
\bauthor{\bsnm{Feng},~\bfnm{Yihao}\binits{Y.}}, \bauthor{\bsnm{Tang},~\bfnm{Ziyang}\binits{Z.}}, \bauthor{\bsnm{Zhang},~\bfnm{Na}\binits{N.}} \AND \bauthor{\bsnm{Liu},~\bfnm{Qiang}\binits{Q.}}
(\byear{2021}).
\btitle{Non-asymptotic confidence intervals of off-policy evaluation: Primal and dual bounds}.
\bjournal{arXiv preprint arXiv:2103.05741}.
\end{barticle}
\endbibitem

\bibitem{ghadimi2013stochastic}
\begin{barticle}[author]
\bauthor{\bsnm{Ghadimi},~\bfnm{Saeed}\binits{S.}} \AND \bauthor{\bsnm{Lan},~\bfnm{Guanghui}\binits{G.}}
(\byear{2013}).
\btitle{Stochastic first-and zeroth-order methods for nonconvex stochastic programming}.
\bjournal{SIAM Journal on Optimization}
\bvolume{23}
\bpages{2341--2368}.
\end{barticle}
\endbibitem

\bibitem{gretton2012kernel}
\begin{barticle}[author]
\bauthor{\bsnm{Gretton},~\bfnm{Arthur}\binits{A.}}, \bauthor{\bsnm{Borgwardt},~\bfnm{Karsten~M}\binits{K.~M.}}, \bauthor{\bsnm{Rasch},~\bfnm{Malte~J}\binits{M.~J.}}, \bauthor{\bsnm{Sch{\"o}lkopf},~\bfnm{Bernhard}\binits{B.}} \AND \bauthor{\bsnm{Smola},~\bfnm{Alexander}\binits{A.}}
(\byear{2012}).
\btitle{A kernel two-sample test}.
\bjournal{The Journal of Machine Learning Research}
\bvolume{13}
\bpages{723--773}.
\end{barticle}
\endbibitem

\bibitem{haarnoja2018softac}
\begin{barticle}[author]
\bauthor{\bsnm{Haarnoja},~\bfnm{Tuomas}\binits{T.}}, \bauthor{\bsnm{Zhou},~\bfnm{Aurick}\binits{A.}}, \bauthor{\bsnm{Hartikainen},~\bfnm{Kristian}\binits{K.}}, \bauthor{\bsnm{Tucker},~\bfnm{George}\binits{G.}}, \bauthor{\bsnm{Ha},~\bfnm{Sehoon}\binits{S.}}, \bauthor{\bsnm{Tan},~\bfnm{Jie}\binits{J.}}, \bauthor{\bsnm{Kumar},~\bfnm{Vikash}\binits{V.}}, \bauthor{\bsnm{Zhu},~\bfnm{Henry}\binits{H.}}, \bauthor{\bsnm{Gupta},~\bfnm{Abhishek}\binits{A.}}, \bauthor{\bsnm{Abbeel},~\bfnm{Pieter}\binits{P.}} \betal{et~al.}
(\byear{2018}).
\btitle{Soft actor-critic algorithms and applications}.
\bjournal{arXiv preprint arXiv:1812.05905}.
\end{barticle}
\endbibitem

\bibitem{hanna2017bootstrapping}
\begin{binproceedings}[author]
\bauthor{\bsnm{Hanna},~\bfnm{Josiah}\binits{J.}}, \bauthor{\bsnm{Stone},~\bfnm{Peter}\binits{P.}} \AND \bauthor{\bsnm{Niekum},~\bfnm{Scott}\binits{S.}}
(\byear{2017}).
\btitle{Bootstrapping with models: Confidence intervals for off-policy evaluation}.
In \bbooktitle{Proceedings of the AAAI Conference on Artificial Intelligence}
\bvolume{31}.
\end{binproceedings}
\endbibitem

\bibitem{hao2021bootstrapping}
\begin{binproceedings}[author]
\bauthor{\bsnm{Hao},~\bfnm{Botao}\binits{B.}}, \bauthor{\bsnm{Ji},~\bfnm{Xiang}\binits{X.}}, \bauthor{\bsnm{Duan},~\bfnm{Yaqi}\binits{Y.}}, \bauthor{\bsnm{Lu},~\bfnm{Hao}\binits{H.}}, \bauthor{\bsnm{Szepesv{\'a}ri},~\bfnm{Csaba}\binits{C.}} \AND \bauthor{\bsnm{Wang},~\bfnm{Mengdi}\binits{M.}}
(\byear{2021}).
\btitle{Bootstrapping fitted q-evaluation for off-policy inference}.
In \bbooktitle{International Conference on Machine Learning}
\bpages{4074--4084}.
\bpublisher{PMLR}.
\end{binproceedings}
\endbibitem

\bibitem{izmailov2018averaging}
\begin{binproceedings}[author]
\bauthor{\bsnm{Izmailov},~\bfnm{Pavel}\binits{P.}}, \bauthor{\bsnm{Podoprikhin},~\bfnm{Dmitrii}\binits{D.}}, \bauthor{\bsnm{Garipov},~\bfnm{Timur}\binits{T.}}, \bauthor{\bsnm{Vetrov},~\bfnm{Dmitry}\binits{D.}} \AND \bauthor{\bsnm{Wilson},~\bfnm{Andrew~Gordon}\binits{A.~G.}}
(\byear{2018}).
\btitle{Averaging weights leads to wider optima and better generalization}.
In \bbooktitle{34th Conference on Uncertainty in Artificial Intelligence 2018, UAI 2018}
\bpages{876--885}.
\bpublisher{Association For Uncertainty in Artificial Intelligence (AUAI)}.
\end{binproceedings}
\endbibitem

\bibitem{jiang2020minimax}
\begin{barticle}[author]
\bauthor{\bsnm{Jiang},~\bfnm{Nan}\binits{N.}} \AND \bauthor{\bsnm{Huang},~\bfnm{Jiawei}\binits{J.}}
(\byear{2020}).
\btitle{Minimax value interval for off-policy evaluation and policy optimization}.
\bjournal{Advances in Neural Information Processing Systems}
\bvolume{33}
\bpages{2747--2758}.
\end{barticle}
\endbibitem

\bibitem{jiang2016doubly}
\begin{binproceedings}[author]
\bauthor{\bsnm{Jiang},~\bfnm{Nan}\binits{N.}} \AND \bauthor{\bsnm{Li},~\bfnm{Lihong}\binits{L.}}
(\byear{2016}).
\btitle{Doubly robust off-policy value evaluation for reinforcement learning}.
In \bbooktitle{International Conference on Machine Learning}
\bpages{652--661}.
\bpublisher{PMLR}.
\end{binproceedings}
\endbibitem

\bibitem{jin2021pessimism}
\begin{binproceedings}[author]
\bauthor{\bsnm{Jin},~\bfnm{Ying}\binits{Y.}}, \bauthor{\bsnm{Yang},~\bfnm{Zhuoran}\binits{Z.}} \AND \bauthor{\bsnm{Wang},~\bfnm{Zhaoran}\binits{Z.}}
(\byear{2021}).
\btitle{Is pessimism provably efficient for offline rl?}
In \bbooktitle{International Conference on Machine Learning}
\bpages{5084--5096}.
\bpublisher{PMLR}.
\end{binproceedings}
\endbibitem

\bibitem{kallus2022doubly}
\begin{binproceedings}[author]
\bauthor{\bsnm{Kallus},~\bfnm{Nathan}\binits{N.}}, \bauthor{\bsnm{Mao},~\bfnm{Xiaojie}\binits{X.}}, \bauthor{\bsnm{Wang},~\bfnm{Kaiwen}\binits{K.}} \AND \bauthor{\bsnm{Zhou},~\bfnm{Zhengyuan}\binits{Z.}}
(\byear{2022}).
\btitle{Doubly robust distributionally robust off-policy evaluation and learning}.
In \bbooktitle{International Conference on Machine Learning}
\bpages{10598--10632}.
\bpublisher{PMLR}.
\end{binproceedings}
\endbibitem

\bibitem{kallus2019intrinsically}
\begin{barticle}[author]
\bauthor{\bsnm{Kallus},~\bfnm{Nathan}\binits{N.}} \AND \bauthor{\bsnm{Uehara},~\bfnm{Masatoshi}\binits{M.}}
(\byear{2019}).
\btitle{Intrinsically efficient, stable, and bounded off-policy evaluation for reinforcement learning}.
\bjournal{Advances in Neural Information Processing Systems}
\bvolume{32}.
\end{barticle}
\endbibitem

\bibitem{kallus2020double}
\begin{barticle}[author]
\bauthor{\bsnm{Kallus},~\bfnm{Nathan}\binits{N.}} \AND \bauthor{\bsnm{Uehara},~\bfnm{Masatoshi}\binits{M.}}
(\byear{2020}).
\btitle{Double reinforcement learning for efficient off-policy evaluation in markov decision processes}.
\bjournal{The Journal of Machine Learning Research}
\bvolume{21}
\bpages{6742--6804}.
\end{barticle}
\endbibitem

\bibitem{kornowski2022oracle}
\begin{barticle}[author]
\bauthor{\bsnm{Kornowski},~\bfnm{Guy}\binits{G.}} \AND \bauthor{\bsnm{Shamir},~\bfnm{Ohad}\binits{O.}}
(\byear{2022}).
\btitle{Oracle Complexity in Nonsmooth Nonconvex Optimization}.
\bjournal{Journal of Machine Learning Research}
\bvolume{23}
\bpages{1--44}.
\end{barticle}
\endbibitem

\bibitem{kumar2019stabilizing}
\begin{barticle}[author]
\bauthor{\bsnm{Kumar},~\bfnm{Aviral}\binits{A.}}, \bauthor{\bsnm{Fu},~\bfnm{Justin}\binits{J.}}, \bauthor{\bsnm{Soh},~\bfnm{Matthew}\binits{M.}}, \bauthor{\bsnm{Tucker},~\bfnm{George}\binits{G.}} \AND \bauthor{\bsnm{Levine},~\bfnm{Sergey}\binits{S.}}
(\byear{2019}).
\btitle{Stabilizing off-policy q-learning via bootstrapping error reduction}.
\bjournal{Advances in Neural Information Processing Systems}
\bvolume{32}.
\end{barticle}
\endbibitem

\bibitem{kumar2020conservative}
\begin{barticle}[author]
\bauthor{\bsnm{Kumar},~\bfnm{Aviral}\binits{A.}}, \bauthor{\bsnm{Zhou},~\bfnm{Aurick}\binits{A.}}, \bauthor{\bsnm{Tucker},~\bfnm{George}\binits{G.}} \AND \bauthor{\bsnm{Levine},~\bfnm{Sergey}\binits{S.}}
(\byear{2020}).
\btitle{Conservative q-learning for offline reinforcement learning}.
\bjournal{Advances in Neural Information Processing Systems}
\bvolume{33}
\bpages{1179--1191}.
\end{barticle}
\endbibitem

\bibitem{le2019batch}
\begin{binproceedings}[author]
\bauthor{\bsnm{Le},~\bfnm{Hoang}\binits{H.}}, \bauthor{\bsnm{Voloshin},~\bfnm{Cameron}\binits{C.}} \AND \bauthor{\bsnm{Yue},~\bfnm{Yisong}\binits{Y.}}
(\byear{2019}).
\btitle{Batch policy learning under constraints}.
In \bbooktitle{International Conference on Machine Learning}
\bpages{3703--3712}.
\bpublisher{PMLR}.
\end{binproceedings}
\endbibitem

\bibitem{levine2020offline}
\begin{barticle}[author]
\bauthor{\bsnm{Levine},~\bfnm{Sergey}\binits{S.}}, \bauthor{\bsnm{Kumar},~\bfnm{Aviral}\binits{A.}}, \bauthor{\bsnm{Tucker},~\bfnm{George}\binits{G.}} \AND \bauthor{\bsnm{Fu},~\bfnm{Justin}\binits{J.}}
(\byear{2020}).
\btitle{Offline reinforcement learning: Tutorial, review, and perspectives on open problems}.
\bjournal{arXiv preprint arXiv:2005.01643}.
\end{barticle}
\endbibitem

\bibitem{li2022reinforcement}
\begin{barticle}[author]
\bauthor{\bsnm{Li},~\bfnm{Mengbing}\binits{M.}}, \bauthor{\bsnm{Shi},~\bfnm{Chengchun}\binits{C.}}, \bauthor{\bsnm{Wu},~\bfnm{Zhenke}\binits{Z.}} \AND \bauthor{\bsnm{Fryzlewicz},~\bfnm{Piotr}\binits{P.}}
(\byear{2022}).
\btitle{Reinforcement Learning in Possibly Nonstationary Environments}.
\bjournal{arXiv preprint arXiv:2203.01707}.
\end{barticle}
\endbibitem

\bibitem{liao2020off}
\begin{barticle}[author]
\bauthor{\bsnm{Liao},~\bfnm{Peng}\binits{P.}}, \bauthor{\bsnm{Klasnja},~\bfnm{Predrag}\binits{P.}} \AND \bauthor{\bsnm{Murphy},~\bfnm{Susan}\binits{S.}}
(\byear{2020}).
\btitle{Off-policy estimation of long-term average outcomes with applications to mobile health}.
\bjournal{Journal of the American Statistical Association}
\bpages{1--10}.
\end{barticle}
\endbibitem

\bibitem{liao2022batch}
\begin{barticle}[author]
\bauthor{\bsnm{Liao},~\bfnm{Peng}\binits{P.}}, \bauthor{\bsnm{Qi},~\bfnm{Zhengling}\binits{Z.}}, \bauthor{\bsnm{Wan},~\bfnm{Runzhe}\binits{R.}}, \bauthor{\bsnm{Klasnja},~\bfnm{Predrag}\binits{P.}} \AND \bauthor{\bsnm{Murphy},~\bfnm{Susan~A}\binits{S.~A.}}
(\byear{2022}).
\btitle{Batch policy learning in average reward markov decision processes}.
\bjournal{The Annals of Statistics}
\bvolume{50}
\bpages{3364--3387}.
\end{barticle}
\endbibitem

\bibitem{liu2018breaking}
\begin{barticle}[author]
\bauthor{\bsnm{Liu},~\bfnm{Qiang}\binits{Q.}}, \bauthor{\bsnm{Li},~\bfnm{Lihong}\binits{L.}}, \bauthor{\bsnm{Tang},~\bfnm{Ziyang}\binits{Z.}} \AND \bauthor{\bsnm{Zhou},~\bfnm{Dengyong}\binits{D.}}
(\byear{2018}).
\btitle{Breaking the curse of horizon: Infinite-horizon off-policy estimation}.
\bjournal{Advances in Neural Information Processing Systems}
\bvolume{31}.
\end{barticle}
\endbibitem

\bibitem{lu2020universal}
\begin{barticle}[author]
\bauthor{\bsnm{Lu},~\bfnm{Yulong}\binits{Y.}} \AND \bauthor{\bsnm{Lu},~\bfnm{Jianfeng}\binits{J.}}
(\byear{2020}).
\btitle{A universal approximation theorem of deep neural networks for expressing probability distributions}.
\bjournal{Advances in Neural Information Processing Systems}
\bvolume{33}
\bpages{3094--3105}.
\end{barticle}
\endbibitem

\bibitem{luckett2020estimating}
\begin{barticle}[author]
\bauthor{\bsnm{Luckett},~\bfnm{Daniel~J}\binits{D.~J.}}, \bauthor{\bsnm{Laber},~\bfnm{Eric~B}\binits{E.~B.}}, \bauthor{\bsnm{Kahkoska},~\bfnm{Anna~R}\binits{A.~R.}}, \bauthor{\bsnm{Maahs},~\bfnm{David~M}\binits{D.~M.}}, \bauthor{\bsnm{Mayer-Davis},~\bfnm{Elizabeth}\binits{E.}} \AND \bauthor{\bsnm{Kosorok},~\bfnm{Michael~R}\binits{M.~R.}}
(\byear{2020}).
\btitle{Estimating dynamic treatment regimes in mobile health using v-learning}.
\bjournal{Journal of the American Statistical Association}
\bvolume{115}
\bpages{692--706}.
\end{barticle}
\endbibitem

\bibitem{mahadevan2014proximal}
\begin{barticle}[author]
\bauthor{\bsnm{Mahadevan},~\bfnm{Sridhar}\binits{S.}}, \bauthor{\bsnm{Liu},~\bfnm{Bo}\binits{B.}}, \bauthor{\bsnm{Thomas},~\bfnm{Philip}\binits{P.}}, \bauthor{\bsnm{Dabney},~\bfnm{Will}\binits{W.}}, \bauthor{\bsnm{Giguere},~\bfnm{Steve}\binits{S.}}, \bauthor{\bsnm{Jacek},~\bfnm{Nicholas}\binits{N.}}, \bauthor{\bsnm{Gemp},~\bfnm{Ian}\binits{I.}} \AND \bauthor{\bsnm{Liu},~\bfnm{Ji}\binits{J.}}
(\byear{2014}).
\btitle{Proximal reinforcement learning: A new theory of sequential decision making in primal-dual spaces}.
\bjournal{arXiv preprint arXiv:1405.6757}.
\end{barticle}
\endbibitem

\bibitem{mandel2014offline}
\begin{binproceedings}[author]
\bauthor{\bsnm{Mandel},~\bfnm{Travis}\binits{T.}}, \bauthor{\bsnm{Liu},~\bfnm{Yun-En}\binits{Y.-E.}}, \bauthor{\bsnm{Levine},~\bfnm{Sergey}\binits{S.}}, \bauthor{\bsnm{Brunskill},~\bfnm{Emma}\binits{E.}} \AND \bauthor{\bsnm{Popovic},~\bfnm{Zoran}\binits{Z.}}
(\byear{2014}).
\btitle{Offline policy evaluation across representations with applications to educational games.}
In \bbooktitle{AAMAS}
\bvolume{1077}.
\end{binproceedings}
\endbibitem

\bibitem{marling2020ohiot1dm}
\begin{barticle}[author]
\bauthor{\bsnm{Marling},~\bfnm{Cindy}\binits{C.}} \AND \bauthor{\bsnm{Bunescu},~\bfnm{Razvan}\binits{R.}}
(\byear{2020}).
\btitle{The ohiot1dm dataset for blood glucose level prediction: Update 2020}.
\bjournal{KHD@ IJCAI}.
\end{barticle}
\endbibitem

\bibitem{mokhtari2020unified}
\begin{binproceedings}[author]
\bauthor{\bsnm{Mokhtari},~\bfnm{Aryan}\binits{A.}}, \bauthor{\bsnm{Ozdaglar},~\bfnm{Asuman}\binits{A.}} \AND \bauthor{\bsnm{Pattathil},~\bfnm{Sarath}\binits{S.}}
(\byear{2020}).
\btitle{A unified analysis of extra-gradient and optimistic gradient methods for saddle point problems: Proximal point approach}.
In \bbooktitle{International Conference on Artificial Intelligence and Statistics}
\bpages{1497--1507}.
\bpublisher{PMLR}.
\end{binproceedings}
\endbibitem

\bibitem{murphy2001marginal}
\begin{barticle}[author]
\bauthor{\bsnm{Murphy},~\bfnm{Susan~A}\binits{S.~A.}}, \bauthor{\bparticle{van~der} \bsnm{Laan},~\bfnm{Mark~J}\binits{M.~J.}}, \bauthor{\bsnm{Robins},~\bfnm{James~M}\binits{J.~M.}} \AND \bauthor{\bsnm{Group},~\bfnm{Conduct Problems Prevention~Research}\binits{C.~P. P.~R.}}
(\byear{2001}).
\btitle{Marginal mean models for dynamic regimes}.
\bjournal{Journal of the American Statistical Association}
\bvolume{96}
\bpages{1410--1423}.
\end{barticle}
\endbibitem

\bibitem{nachum2020reinforcement}
\begin{barticle}[author]
\bauthor{\bsnm{Nachum},~\bfnm{Ofir}\binits{O.}} \AND \bauthor{\bsnm{Dai},~\bfnm{Bo}\binits{B.}}
(\byear{2021}).
\btitle{Reinforcement learning via fenchel-rockafellar duality}.
\bjournal{arXiv preprint arXiv:2001.01866}.
\end{barticle}
\endbibitem

\bibitem{parikh2014proximal}
\begin{barticle}[author]
\bauthor{\bsnm{Parikh},~\bfnm{Neal}\binits{N.}}, \bauthor{\bsnm{Boyd},~\bfnm{Stephen}\binits{S.}} \betal{et~al.}
(\byear{2014}).
\btitle{Proximal algorithms}.
\bjournal{Foundations and trends{\textregistered} in Optimization}
\bvolume{1}
\bpages{127--239}.
\end{barticle}
\endbibitem

\bibitem{pinelis2012optimum}
\begin{barticle}[author]
\bauthor{\bsnm{Pinelis},~\bfnm{Iosif}\binits{I.}}
(\byear{2012}).
\btitle{Optimum bounds for the distributions of martingales in Banach spaces}.
\bjournal{arXiv preprint arXiv:1208.2200}.
\end{barticle}
\endbibitem

\bibitem{precup2000eligibility}
\begin{barticle}[author]
\bauthor{\bsnm{Precup},~\bfnm{Doina}\binits{D.}}
(\byear{2000}).
\btitle{Eligibility traces for off-policy policy evaluation}.
\bjournal{Computer Science Department Faculty Publication Series}
\bpages{80}.
\end{barticle}
\endbibitem

\bibitem{rahimi2007random}
\begin{barticle}[author]
\bauthor{\bsnm{Rahimi},~\bfnm{Ali}\binits{A.}} \AND \bauthor{\bsnm{Recht},~\bfnm{Benjamin}\binits{B.}}
(\byear{2007}).
\btitle{Random features for large-scale kernel machines}.
\bjournal{Advances in Neural Information Processing Systems}
\bvolume{20}.
\end{barticle}
\endbibitem

\bibitem{ramprasad2022online}
\begin{barticle}[author]
\bauthor{\bsnm{Ramprasad},~\bfnm{Pratik}\binits{P.}}, \bauthor{\bsnm{Li},~\bfnm{Yuantong}\binits{Y.}}, \bauthor{\bsnm{Yang},~\bfnm{Zhuoran}\binits{Z.}}, \bauthor{\bsnm{Wang},~\bfnm{Zhaoran}\binits{Z.}}, \bauthor{\bsnm{Sun},~\bfnm{Will~Wei}\binits{W.~W.}} \AND \bauthor{\bsnm{Cheng},~\bfnm{Guang}\binits{G.}}
(\byear{2022}).
\btitle{Online bootstrap inference for policy evaluation in reinforcement learning}.
\bjournal{Journal of the American Statistical Association}
\bpages{1--14}.
\end{barticle}
\endbibitem

\bibitem{reddi2016stochastic}
\begin{binproceedings}[author]
\bauthor{\bsnm{Reddi},~\bfnm{Sashank~J}\binits{S.~J.}}, \bauthor{\bsnm{Hefny},~\bfnm{Ahmed}\binits{A.}}, \bauthor{\bsnm{Sra},~\bfnm{Suvrit}\binits{S.}}, \bauthor{\bsnm{Poczos},~\bfnm{Barnabas}\binits{B.}} \AND \bauthor{\bsnm{Smola},~\bfnm{Alex}\binits{A.}}
(\byear{2016}).
\btitle{Stochastic variance reduction for nonconvex optimization}.
In \bbooktitle{International conference on machine learning}
\bpages{314--323}.
\bpublisher{PMLR}.
\end{binproceedings}
\endbibitem

\bibitem{reem2019re}
\begin{barticle}[author]
\bauthor{\bsnm{Reem},~\bfnm{Daniel}\binits{D.}}, \bauthor{\bsnm{Reich},~\bfnm{Simeon}\binits{S.}} \AND \bauthor{\bsnm{De~Pierro},~\bfnm{Alvaro}\binits{A.}}
(\byear{2019}).
\btitle{Re-examination of Bregman functions and new properties of their divergences}.
\bjournal{Optimization}
\bvolume{68}
\bpages{279--348}.
\end{barticle}
\endbibitem

\bibitem{renyi1961measures}
\begin{binproceedings}[author]
\bauthor{\bsnm{R{\'e}nyi},~\bfnm{Alfr{\'e}d}\binits{A.}}
(\byear{1961}).
\btitle{On measures of entropy and information}.
In \bbooktitle{Proceedings of the Fourth Berkeley Symposium on Mathematical Statistics and Probability, Volume 1: Contributions to the Theory of Statistics}
\bvolume{4}
\bpages{547--562}.
\bpublisher{University of California Press}.
\end{binproceedings}
\endbibitem

\bibitem{rodbard2009interpretation}
\begin{barticle}[author]
\bauthor{\bsnm{Rodbard},~\bfnm{David}\binits{D.}}
(\byear{2009}).
\btitle{Interpretation of continuous glucose monitoring data: glycemic variability and quality of glycemic control}.
\bjournal{Diabetes Technology \& Therapeutics}
\bvolume{11}
\bpages{S--55}.
\end{barticle}
\endbibitem

\bibitem{shi2022minimax}
\begin{binproceedings}[author]
\bauthor{\bsnm{Shi},~\bfnm{Chengchun}\binits{C.}}, \bauthor{\bsnm{Uehara},~\bfnm{Masatoshi}\binits{M.}}, \bauthor{\bsnm{Huang},~\bfnm{Jiawei}\binits{J.}} \AND \bauthor{\bsnm{Jiang},~\bfnm{Nan}\binits{N.}}
(\byear{2022}).
\btitle{A minimax learning approach to off-policy evaluation in confounded partially observable markov decision processes}.
In \bbooktitle{International Conference on Machine Learning}
\bpages{20057--20094}.
\bpublisher{PMLR}.
\end{binproceedings}
\endbibitem

\bibitem{shi2021deeply}
\begin{binproceedings}[author]
\bauthor{\bsnm{Shi},~\bfnm{Chengchun}\binits{C.}}, \bauthor{\bsnm{Wan},~\bfnm{Runzhe}\binits{R.}}, \bauthor{\bsnm{Chernozhukov},~\bfnm{Victor}\binits{V.}} \AND \bauthor{\bsnm{Song},~\bfnm{Rui}\binits{R.}}
(\byear{2021}).
\btitle{Deeply-debiased off-policy interval estimation}.
In \bbooktitle{International Conference on Machine Learning}
\bpages{9580--9591}.
\bpublisher{PMLR}.
\end{binproceedings}
\endbibitem

\bibitem{shi2020statistical}
\begin{barticle}[author]
\bauthor{\bsnm{Shi},~\bfnm{Chengchun}\binits{C.}}, \bauthor{\bsnm{Zhang},~\bfnm{Sheng}\binits{S.}}, \bauthor{\bsnm{Lu},~\bfnm{Wenbin}\binits{W.}} \AND \bauthor{\bsnm{Song},~\bfnm{Rui}\binits{R.}}
(\byear{2020}).
\btitle{Statistical inference of the value function for reinforcement learning in infinite horizon settings}.
\bjournal{arXiv preprint arXiv:2001.04515}.
\end{barticle}
\endbibitem

\bibitem{shi2022off}
\begin{barticle}[author]
\bauthor{\bsnm{Shi},~\bfnm{Chengchun}\binits{C.}}, \bauthor{\bsnm{Zhu},~\bfnm{Jin}\binits{J.}}, \bauthor{\bsnm{Ye},~\bfnm{Shen}\binits{S.}}, \bauthor{\bsnm{Luo},~\bfnm{Shikai}\binits{S.}}, \bauthor{\bsnm{Zhu},~\bfnm{Hongtu}\binits{H.}} \AND \bauthor{\bsnm{Song},~\bfnm{Rui}\binits{R.}}
(\byear{2022}).
\btitle{Off-policy confidence interval estimation with confounded markov decision process}.
\bjournal{Journal of the American Statistical Association}
\bpages{1--12}.
\end{barticle}
\endbibitem

\bibitem{sriperumbudur2011universality}
\begin{barticle}[author]
\bauthor{\bsnm{Sriperumbudur},~\bfnm{Bharath~K}\binits{B.~K.}}, \bauthor{\bsnm{Fukumizu},~\bfnm{Kenji}\binits{K.}} \AND \bauthor{\bsnm{Lanckriet},~\bfnm{Gert~RG}\binits{G.~R.}}
(\byear{2011}).
\btitle{Universality, Characteristic Kernels and RKHS Embedding of Measures.}
\bjournal{Journal of Machine Learning Research}
\bvolume{12}.
\end{barticle}
\endbibitem

\bibitem{sutton2018reinforcement}
\begin{bbook}[author]
\bauthor{\bsnm{Sutton},~\bfnm{Richard~S}\binits{R.~S.}} \AND \bauthor{\bsnm{Barto},~\bfnm{Andrew~G}\binits{A.~G.}}
(\byear{2018}).
\btitle{Reinforcement learning: An introduction}.
\bpublisher{MIT press}.
\end{bbook}
\endbibitem

\bibitem{theocharous2020reinforcement}
\begin{barticle}[author]
\bauthor{\bsnm{Theocharous},~\bfnm{Georgios}\binits{G.}}, \bauthor{\bsnm{Chandak},~\bfnm{Yash}\binits{Y.}}, \bauthor{\bsnm{Thomas},~\bfnm{Philip~S}\binits{P.~S.}} \AND \bauthor{\bparticle{de} \bsnm{Nijs},~\bfnm{Frits}\binits{F.}}
(\byear{2020}).
\btitle{Reinforcement learning for strategic recommendations}.
\bjournal{arXiv preprint arXiv:2009.07346}.
\end{barticle}
\endbibitem

\bibitem{thomas2016data}
\begin{binproceedings}[author]
\bauthor{\bsnm{Thomas},~\bfnm{Philip}\binits{P.}} \AND \bauthor{\bsnm{Brunskill},~\bfnm{Emma}\binits{E.}}
(\byear{2016}).
\btitle{Data-efficient off-policy policy evaluation for reinforcement learning}.
In \bbooktitle{International Conference on Machine Learning}
\bpages{2139--2148}.
\bpublisher{PMLR}.
\end{binproceedings}
\endbibitem

\bibitem{thomas2015high}
\begin{binproceedings}[author]
\bauthor{\bsnm{Thomas},~\bfnm{Philip}\binits{P.}}, \bauthor{\bsnm{Theocharous},~\bfnm{Georgios}\binits{G.}} \AND \bauthor{\bsnm{Ghavamzadeh},~\bfnm{Mohammad}\binits{M.}}
(\byear{2015}).
\btitle{High-confidence off-policy evaluation}.
In \bbooktitle{Proceedings of the AAAI Conference on Artificial Intelligence}
\bvolume{29}.
\end{binproceedings}
\endbibitem

\bibitem{tropp2011freedman}
\begin{barticle}[author]
\bauthor{\bsnm{Tropp},~\bfnm{Joel~A}\binits{J.~A.}}
(\byear{2011}).
\btitle{Freedman’s Inequality for Matrix Martingales}.
\bjournal{Electronic Communications in Probability}
\bvolume{16}
\bpages{262--270}.
\end{barticle}
\endbibitem

\bibitem{tsybakov2009introduction}
\begin{barticle}[author]
\bauthor{\bsnm{Tsybakov},~\bfnm{Alexandre~B}\binits{A.~B.}}
(\byear{2004}).
\btitle{Introduction to nonparametric estimation, 2009}.
\bjournal{Springer Series in Statistics}
\bvolume{9}.
\end{barticle}
\endbibitem

\bibitem{uehara2020minimax}
\begin{binproceedings}[author]
\bauthor{\bsnm{Uehara},~\bfnm{Masatoshi}\binits{M.}}, \bauthor{\bsnm{Huang},~\bfnm{Jiawei}\binits{J.}} \AND \bauthor{\bsnm{Jiang},~\bfnm{Nan}\binits{N.}}
(\byear{2020}).
\btitle{Minimax weight and q-function learning for off-policy evaluation}.
In \bbooktitle{International Conference on Machine Learning}
\bpages{9659--9668}.
\bpublisher{PMLR}.
\end{binproceedings}
\endbibitem

\bibitem{uehara2022finite}
\begin{barticle}[author]
\bauthor{\bsnm{Uehara},~\bfnm{Masatoshi}\binits{M.}}, \bauthor{\bsnm{Imaizumi},~\bfnm{Masaaki}\binits{M.}}, \bauthor{\bsnm{Jiang},~\bfnm{Nan}\binits{N.}}, \bauthor{\bsnm{Kallus},~\bfnm{Nathan}\binits{N.}}, \bauthor{\bsnm{Sun},~\bfnm{Wen}\binits{W.}} \AND \bauthor{\bsnm{Xie},~\bfnm{Tengyang}\binits{T.}}
(\byear{2022}).
\btitle{Finite sample analysis of minimax offline reinforcement learning: Completeness, fast rates and first-order efficiency}.
\bjournal{arXiv preprint arXiv:2102.02981}.
\end{barticle}
\endbibitem

\bibitem{uehara2022review}
\begin{barticle}[author]
\bauthor{\bsnm{Uehara},~\bfnm{Masatoshi}\binits{M.}}, \bauthor{\bsnm{Shi},~\bfnm{Chengchun}\binits{C.}} \AND \bauthor{\bsnm{Kallus},~\bfnm{Nathan}\binits{N.}}
(\byear{2022}).
\btitle{A Review of Off-Policy Evaluation in Reinforcement Learning}.
\bjournal{arXiv preprint arXiv:2212.06355}.
\end{barticle}
\endbibitem

\bibitem{wang2020statistical}
\begin{barticle}[author]
\bauthor{\bsnm{Wang},~\bfnm{Ruosong}\binits{R.}}, \bauthor{\bsnm{Foster},~\bfnm{Dean~P}\binits{D.~P.}} \AND \bauthor{\bsnm{Kakade},~\bfnm{Sham~M}\binits{S.~M.}}
(\byear{2020}).
\btitle{What are the statistical limits of offline RL with linear function approximation?}
\bjournal{arXiv preprint arXiv:2010.11895}.
\end{barticle}
\endbibitem

\bibitem{xie2021bellman}
\begin{barticle}[author]
\bauthor{\bsnm{Xie},~\bfnm{Tengyang}\binits{T.}}, \bauthor{\bsnm{Cheng},~\bfnm{Ching-An}\binits{C.-A.}}, \bauthor{\bsnm{Jiang},~\bfnm{Nan}\binits{N.}}, \bauthor{\bsnm{Mineiro},~\bfnm{Paul}\binits{P.}} \AND \bauthor{\bsnm{Agarwal},~\bfnm{Alekh}\binits{A.}}
(\byear{2021}).
\btitle{Bellman-consistent pessimism for offline reinforcement learning}.
\bjournal{Advances in neural information processing systems}
\bvolume{34}.
\end{barticle}
\endbibitem

\bibitem{xie2019towards}
\begin{barticle}[author]
\bauthor{\bsnm{Xie},~\bfnm{Tengyang}\binits{T.}}, \bauthor{\bsnm{Ma},~\bfnm{Yifei}\binits{Y.}} \AND \bauthor{\bsnm{Wang},~\bfnm{Yu-Xiang}\binits{Y.-X.}}
(\byear{2019}).
\btitle{Towards optimal off-policy evaluation for reinforcement learning with marginalized importance sampling}.
\bjournal{Advances in Neural Information Processing Systems}
\bvolume{32}.
\end{barticle}
\endbibitem

\bibitem{zhang2020gendice}
\begin{barticle}[author]
\bauthor{\bsnm{Zhang},~\bfnm{Ruiyi}\binits{R.}}, \bauthor{\bsnm{Dai},~\bfnm{Bo}\binits{B.}}, \bauthor{\bsnm{Li},~\bfnm{Lihong}\binits{L.}} \AND \bauthor{\bsnm{Schuurmans},~\bfnm{Dale}\binits{D.}}
(\byear{2020}).
\btitle{Gendice: Generalized offline estimation of stationary values}.
\bjournal{arXiv preprint arXiv:2002.09072}.
\end{barticle}
\endbibitem

\bibitem{zhou2022estimating}
\begin{barticle}[author]
\bauthor{\bsnm{Zhou},~\bfnm{Wenzhuo}\binits{W.}}, \bauthor{\bsnm{Zhu},~\bfnm{Ruoqing}\binits{R.}} \AND \bauthor{\bsnm{Qu},~\bfnm{Annie}\binits{A.}}
(\byear{2022}).
\btitle{Estimating optimal infinite horizon dynamic treatment regimes via pt-learning}.
\bjournal{Journal of the American Statistical Association}
\bpages{1--14}.
\end{barticle}
\endbibitem

\bibitem{zhu2020causal}
\begin{binproceedings}[author]
\bauthor{\bsnm{Zhu},~\bfnm{Liangyu}\binits{L.}}, \bauthor{\bsnm{Lu},~\bfnm{Wenbin}\binits{W.}} \AND \bauthor{\bsnm{Song},~\bfnm{Rui}\binits{R.}}
(\byear{2020}).
\btitle{Causal Effect Estimation and Optimal Dose Suggestions in Mobile Health}.
In \bbooktitle{International Conference on Machine Learning}
\bpages{11588--11598}.
\bpublisher{PMLR}.
\end{binproceedings}
\endbibitem

\end{thebibliography}


\setcounter{tocdepth}{-1}

\newpage 

\begin{center} 
{\textbf{SUPPLEMENTARY MATERIALS FOR\\ ``DISTRIBUTIONAL SHIFT-AWARE OFF-POLICY INTERVAL ESTIMATION: \\ A UNIFIED ERROR QUANTIFICATION FRAMEWORK''}}
\end{center}

\begin{center} 
\normalsize{Wenzhuo Zhou, Yuhan Li, Ruoqing Zhu, and Annie Qu}
\end{center}

\bigskip

\noindent \textbf{A} \quad \textbf{Interval Validity and Tightness for Bias Quantification} \hfill {\color{blue}\textbf{2}}

\smallskip 

\noindent \textbf{B} \quad \textbf{Expressivity and Role of Q-function Class} 
\hfill {\color{blue}\textbf{2}}

\smallskip 

\noindent \textbf{C} \quad \textbf{Probabilistic Tools in Statistical Learning} 
\hfill {\color{blue}\textbf{3}}

\smallskip 

\noindent \textbf{D} \quad \textbf{Details of Numerical Experiments and Additional Results} 
\hfill {\color{blue}\textbf{4}}

\smallskip 

\noindent \textbf{E} \quad \textbf{Technical Proofs and Additional Theoretical Results} 
\hfill {\color{blue}\textbf{6}}

E.1 \quad Proof of Lemma {\color{blue}3.1} \hfill {\color{blue}6} 

E.2 \quad Proof of Theorem {\color{blue}A.1} \hfill {\color{blue}7} 

E.3 \quad Proof of Corollary {\color{blue}A.1} \hfill {\color{blue}8} 

E.4 \quad Proof of Lemma  {\color{blue}E.1} \hfill {\color{blue}8} 

E.5 \quad Proof of Theorem   {\color{blue}3.1} \hfill {\color{blue}9} 

E.6 \quad Proof of Theorem   {\color{blue}B.1} \hfill {\color{blue}10} 

E.7 \quad Proof of Lemma  {\color{blue}3.2} \hfill {\color{blue}12} 

E.8 \quad Proof of Theorem {\color{blue}3.2} \hfill {\color{blue}12} 

E.9 \quad Proof of Theorem {\color{blue}3.3} \hfill {\color{blue}14} 

E.10 \; Proof of Corollary {\color{blue}3.1} \hfill {\color{blue}17} 

E.11 \; Proof of Corollary {\color{blue}4.1} \hfill {\color{blue}19} 

E.12 \; Proof of Theorem {\color{blue}4.1} \hfill {\color{blue}21} 

E.13 \; Proof of Theorem {\color{blue}3.4} \hfill {\color{blue}23} 

E.14 \; Proof of Proposition {\color{blue}4.1} \hfill {\color{blue}25} 

E.15 \; Proof of Lemma {\color{blue}4.1} \hfill {\color{blue}26} 

E.16 \; Proof of Theorem {\color{blue}4.2} \hfill {\color{blue}28} 

E.17 \; Proof of Theorem  {\color{blue}6.1} \hfill {\color{blue}30} 

E.18 \; Proof of Theorem  {\color{blue}6.2} \hfill {\color{blue}33} 

E.19 \; Proof of Lemma  {\color{blue}E.2} \hfill {\color{blue}35} 

E.20 \; Proof of Theorem {\color{blue}6.3} \hfill {\color{blue}36} 

\quad E.20.1 \quad Proof of the validity of neutral CI \hfill {\color{blue}36} 

\quad E.20.2 \quad Proof of the error-robust finite sample bound for neutral CI \hfill {\color{blue}37} 

E.21 \; Proof of Theorem {\color{blue}6.4} \hfill {\color{blue}40} 

\quad E.21.1 \quad Proof of the validity of pessimist \hfill {\color{blue}40} 

\quad E.21.2 \quad Proof of the error-robust finite sample bound for pessimistic CI \hfill {\color{blue}41} 

E.22 \; Proof of Theorem {\color{blue}E.3} \hfill {\color{blue}43} 

E.23 \; Proof of Theorem {\color{blue}E.4} \hfill {\color{blue}44} 

E.24 \; Proof of Theorem {\color{blue}6.5} \hfill {\color{blue}47} 

\smallskip 

\noindent \textbf{Reference} \hfill {\color{blue}50} 

\end{document}